%% file: neurips_2026.tex
\documentclass{article}

\PassOptionsToPackage{numbers, compress}{natbib}
\usepackage[preprint]{neurips_2026}

\usepackage[utf8]{inputenc} % allow utf-8 input
\usepackage[T1]{fontenc}    % use 8-bit T1 fonts
\usepackage{hyperref}       % hyperlinks
\usepackage{url}            % simple URL typesetting
\usepackage{booktabs}       % professional-quality tables
\usepackage{amsfonts}       % blackboard math symbols
\usepackage{nicefrac}       % compact symbols for 1/2, etc.
\usepackage{microtype}      % microtypography
\usepackage{xcolor}         % colors
\usepackage{amsmath}
\usepackage{multirow}
\usepackage{graphicx}
\usepackage{makecell}

\input{preamable}
\title{CL-CLIP: CLIP-Based Continual Learning Framework with Cost-Volume Category Decoupling for Object Detection}

\author{%
\textbf{Zihan Liu\textsuperscript{1}\footnotemark[1]}, 
\textbf{Yuguang Yang\textsuperscript{3,2}\footnotemark[1]}, 
\textbf{Shengjie Su\textsuperscript{4}\footnotemark[1]}, 
\textbf{Jianing Pang\textsuperscript{5}}, 
\textbf{Linlin Yang\textsuperscript{6}}\footnotemark[2], \\
\textbf{Chunyu Xie\textsuperscript{2}}\footnotemark[2], 
\textbf{Nikolai Yu. Zolotykh\textsuperscript{7}\footnotemark[3]}, 
\textbf{Baochang Zhang\textsuperscript{8}\footnotemark[3]},\\
\\
\textsuperscript{1}National College for Excellent Engineers, Beihang University \\
\textsuperscript{2}360 AI Research, Qihoo 360 \\
\textsuperscript{3}School of Electronic Information Engineering, Beihang University \\
\textsuperscript{4}School of Cyber Science and Technology, Beihang University \\
\textsuperscript{5}School of Computer Science and Engineering, Beihang University \\
\textsuperscript{6}State Key Laboratory of Media Convergence and Communication, \\
~~Communication University of China \\
\textsuperscript{7}Institute of Information Technology, Mathematics and Mechanics, \\
~~Lobachebsky University \\
\textsuperscript{8}School of Artificial Intelligence, Beihang University \\
}

\begin{document}

\maketitle

% 这里要改！⬇️Å
\renewcommand{\thefootnote}{\fnsymbol{footnote}}
\footnotetext[1]{Equal contribution.\{zy2523383, guangbuaa, 23371363\}@buaa.edu.cn}
\footnotetext[2]{Corresponding author. \texttt{lyang@cuc.edu.cn}, \texttt{xiechunyu@360.cn}}
% \footnotetext[2]{Corresponding author.\texttt{bczhang@buaa.edu.cn}}
\footnotetext[3]{Project leader.}

\input{sec/Abstract}

\input{sec/Introduction}

\input{sec/Related_works}

\input{sec/Methodology}

\input{sec/Experiments}

{
\small

% [1] Alexander, J.A.\ \& Mozer, M.C.\ (1995) Template-based algorithms for
% connectionist rule extraction. In G.\ Tesauro, D.S.\ Touretzky and T.K.\ Leen
% (eds.), {\it Advances in Neural Information Processing Systems 7},
% pp.\ 609--616. Cambridge, MA: MIT Press.

% [2] Bower, J.M.\ \& Beeman, D.\ (1995) {\it The Book of GENESIS: Exploring
%   Realistic Neural Models with the GEneral NEural SImulation System.}  New York:
% TELOS/Springer--Verlag.

% [3] Hasselmo, M.E., Schnell, E.\ \& Barkai, E.\ (1995) Dynamics of learning and
% recall at excitatory recurrent synapses and cholinergic modulation in rat
% hippocampal region CA3. {\it Journal of Neuroscience} {\bf 15}(7):5249-5262.

\bibliographystyle{plainnat}
\bibliography{references}

}

%%%%%%%%%%%%%%%%%%%%%%%%%%%%%%%%%%%%%%%%%%%%%%%%%%%%%%%%%%%%
\newpage
\input{sec/appendix}

%%%%%%%%%%%%%%%%%%%%%%%%%%%%%%%%%%%%%%%%%%%%%%%%%%%%%%%%%%%%

\newpage
\input{checklist.tex}

\end{document}

%% file: sec/Abstract.tex
\begin{abstract}
  Continual Object Detection (COD) requires a detector to acquire new categories over time while preserving previously learned ones. This goal is closely related to open-vocabulary detection, since both settings require reasoning over categories that are not fully covered by the annotations available at the current training stage. Recent CLIP-based open-vocabulary detectors have shown strong zero-shot generalization, and frameworks such as F-ViT demonstrate that vision-language pretraining can provide powerful zero-shot detection ability for unseen categories. However, real-world deployments cannot remain purely zero-shot: once these detectors are continually updated on newly introduced categories, they suffer severe catastrophic forgetting and quickly lose their previously calibrated detection ability. We therefore propose CL-CLIP, a CLIP-based COD framework that equips open-vocabulary detectors with better continual learning ability through cost-volume-guided category decoupling. Specifically, following CAT-Seg, we compute a CLIP image-text similarity cost volume, defined as dense category-wise response maps between visual tokens and class text embeddings. This zero-shot spatial prior decomposes shared region features into class-specific pathways, which are then processed by a Multi-Expert RoI head. Extensive experiments on PASCAL VOC and MS-COCO show that CL-CLIP substantially improves the F-ViT baseline under continual fine-tuning and achieves  competitive performance with existing continual object detectors, especially in adapting to newly introduced categories while preserving competitive base-class performance.
\end{abstract}

%% file: sec/Introduction.tex
\section{Introduction}
Continual Object Detection (COD) aims to learn novel categories from a task stream while preserving previously acquired detection ability \cite{li2017learning,rebuffi2017icarl,peng2020faster,feng2022overcoming}. Unlike image classification, however, COD must handle multi-object scenes in which old and new categories co-occur spatially. This leads to the background relegation problem. When the model is updated on a new task, previously learned objects appear without annotations and are treated as background, rapidly eroding old-class decision boundaries and causing severe catastrophic forgetting.

Existing COD methods address this problem through knowledge distillation \cite{peng2020faster,feng2022overcoming,huang2023incremental,wang2025gcd}, regularization \cite{kirkpatrick2017overcoming,wu2025demystifying,luo2025gradient}, replay \cite{gupta2022ow,liu2023augmented,an2025ior}, and pseudo-labeling \cite{yang2023pseudo}. These studies provide important mechanisms for preserving previously learned categories during continual learning. In this work, we explore a different route by using open-vocabulary detection (OVD) as a starting point for COD. This route is motivated by the overlap between the two settings, which both require detectors to reason about categories that are not fully covered by the annotations available at a given training stage.

CLIP-based OVD detectors are especially relevant to this route because CLIP provides strong zero-shot ability, which has been widely used to detect categories beyond the training label set. Fine-grained CLIP variants \cite{zhong2022regionclip,jing2024fineclip,xie2025fg} and frameworks such as F-ViT \cite{kuo2022f} show that strong vision-language alignment can provide powerful zero-shot detection ability for unseen categories. However, OVD and COD differ in a crucial way. OVD relies on zero-shot transfer without updating the detector, whereas COD repeatedly fine-tunes it as new categories arrive. Once F-ViT-style frameworks are placed in this continual setting, their shared trainable heads are easily overwritten, and the old-class detection ability quickly deteriorates. Traditional remedies such as distillation or replay do not resolve this shared-head interference, and existing architecture-specific continual designs based on learnable prompts \cite{yi2025idpa} or memory networks \cite{bhatt2024preventing} are typically developed for DETR \cite{carion2020end} or Grounding DINO \cite{liu2024grounding} pipelines rather than for CLIP-based detectors.

To address this limitation, we propose CL-CLIP, a continual object detection framework built on category decoupling for CLIP-based open-vocabulary detectors. Inspired by CAT-Seg \cite{cho2024cat}, which constructs an image-text similarity cost volume to produce category-wise spatial responses for semantic segmentation, we ask a natural question: can such responses serve as a spatial prior for previously learned classes during continual learning, thereby reducing forgetting under incomplete current-task annotations? Following this insight, CL-CLIP converts the category decomposition ability of CLIP into class-specific detection pathways, reducing direct competition between old and new categories during continual updates. 

The underlying principle is that a frozen CLIP model produces category-indexed spatial responses invariant to downstream fine-tuning, serving as a \emph{structurally stable routing signal} that decomposes shared features into class-specific pathways without being corrupted by missing-label supervision on subsequent tasks. This principle is instantiated through two core designs:
\begin{itemize}
    \item \textbf{Cost-volume-guided category decoupling.} The CLIP cost volume provides a zero-shot spatial prior that separates shared region features into per-category pathways, keeping category responses well separated across tasks.
    \item \textbf{Multi-Expert RoI head with drift stabilization.} Each category is assigned a convolutional expert that processes its decoupled features and produces CLIP-aligned detection scores via per-class sigmoid classifiers. Combined with drift regularization on the shared FPN modules, this design reduces cross-class interference and better preserves old-class representations during continual updates.
\end{itemize}
To our knowledge, CL-CLIP is the first framework to systematically adapt cost-volume-based category decomposition for continual object detection, establishing a previously unexplored connection between open-vocabulary spatial priors and architectural anti-forgetting mechanisms. Through extensive experiments on PASCAL VOC \cite{everingham2010pascal} and MS-COCO \cite{lin2014microsoft}, including a systematic benchmark for forgetting across multiple CLIP variants within the F-ViT baseline, we show that CL-CLIP substantially improves continual detection performance and achieves competitive results with existing continual object detectors, with particularly strong adaptation to newly introduced categories.

%% file: sec/Related_works.tex
\section{Related works}
\noindent \textbf{Continual Learning.} Continual learning studies how a model absorbs new knowledge from a task stream without forgetting previous tasks \cite{li2017learning,rebuffi2017icarl}. Our work focuses on continual object detection, where the challenge is not only catastrophic forgetting but also preserving spatially localized representations while reducing region-level interference between old and new categories.

\noindent \textbf{Continual Object Detection.} Existing COD methods mitigate forgetting through several lines of solutions. Distillation methods (Faster-ILOD \cite{peng2020faster}, GCD \cite{wang2025gcd}, MMA \cite{cermelli2022modeling}) transfer old-task knowledge to the current model. Regularization methods (IncDet \cite{liu2020incdet}, GDA \cite{luo2025gradient}, NSGP-RePRE \cite{wu2025demystifying}) constrain parameters or gradients to reduce harmful updates. Replay methods (ABR \cite{liu2023augmented}, IOR \cite{an2025ior}) preserve previous knowledge by reusing stored or generated samples. Pseudo-labeling methods \cite{yang2023pseudo} recover old objects that reappear without annotations. More recent methods such as ROSETTA \cite{yang2022continual} and TARL \cite{zhang2024learning} isolate task-specific parameters to reduce interference. These approaches can be characterized along two axes: \emph{isolation granularity} (task-level prompts or branches vs.\ category-level pathways) and \emph{old-class recovery strategy} (pseudo-labeling, replay or architectural separation). Most prior work operates at the task level and relies on explicit old-class recovery mechanisms. In contrast, CL-CLIP performs category-level decoupling by constructing class-specific RoI pathways processed by per-category experts, requiring neither pseudo-labels nor replay and eliminating the need for task identity at inference.

\noindent \textbf{CLIP Models for Dense Prediction and Continual Adaptation.} CLIP \cite{radford2021learning} has become a strong foundation for open-vocabulary dense prediction. Prior works developed improved or fine-grained CLIP variants such as RegionCLIP, SigLIP2, EVA-CLIP, FineCLIP, and FG-CLIP \cite{zhong2022regionclip,tschannen2025siglip,sun2023eva,jing2024fineclip,xie2025fg}, and frameworks such as F-ViT \cite{kuo2022f} show that a frozen vision-language backbone can support open-vocabulary detection. However, under sequential fine-tuning, F-ViT's shared detection head overwrites representations calibrated for old classes, motivating an explicitly continual architecture.

Recent studies have begun to explore CLIP for continual detection. IODC \cite{huang2023incremental} introduces CLIP through pseudo-labeling and prompt tuning, but depends on pseudo-label quality. TARL \cite{zhang2024learning} applies task-aware parameter isolation on GLIP \cite{li2022grounded} but operates at the task level and relies on a grounding-based backbone. In contrast, CL-CLIP avoids pseudo-labeling and decouples at the category level, enabling per-class pathways without requiring task identity at inference. Inspired by CAT-Seg \cite{cho2024cat}, designed for static open-vocabulary segmentation, we adopt its category-indexed spatial responses as a structural prior for continual detection, removing its cross-class aggregation module to avoid inter-class mixing under missing-label supervision.

%% file: sec/Methodology.tex
\section{Methodology} \label{sec:methodology}

\begin{figure*}[t]
     \centering
     \includegraphics[width=\textwidth]{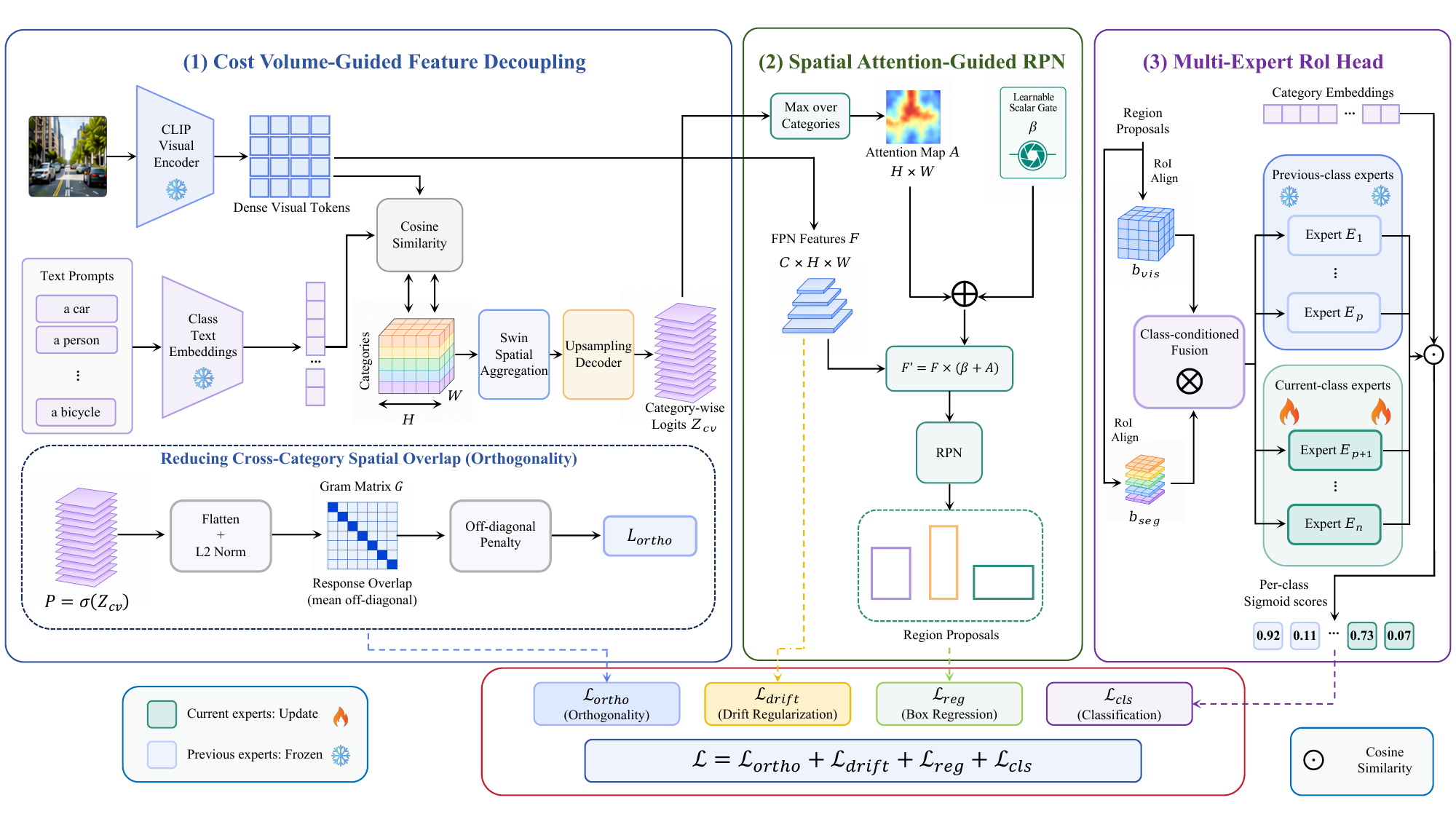}
     \caption{Overview of CL-CLIP. (1) Cost Volume-Guided Feature Decoupling constructs category-wise spatial responses from CLIP image-text similarity;
 (2) Spatial Attention-Guided RPN uses the max-pooled response as a foreground prior; (3) Multi-Expert RoI Head assigns per-category convolutional 
experts with frozen/trainable separation. The orthogonality loss reduces cross-category spatial overlap via a Gram matrix penalty.}
     \label{fig:architecture}
 \end{figure*}

We present CL-CLIP, a framework that equips a CLIP-based open-vocabulary detector with continual learning ability. Starting from a frozen CLIP backbone and the F-ViT detection pipeline, we construct a Similarity Cost Volume from image-text alignment as a zero-shot spatial prior. This prior separates shared multi-scale features into class-specific pathways during sequential fine-tuning. On top of these decoupled features, we build a spatial-attention-guided RPN and a Multi-Expert RoI head to reduce cross-class interference. Throughout the framework, only the CLIP encoders and previous-task experts in the RoI head are frozen; all other modules (cost-volume pathway, FPN, RPN, current-task experts) are trainable. The overall architecture is illustrated in Fig.~\ref{fig:architecture}.

\textbf{Notations}\label{notations}: Let $I$ denote the input image and $\mathcal{F}(\cdot)$ represent the frozen CLIP vision backbone. For a given proposal $b$ and a candidate category set $\mathcal{C}$, CL-CLIP produces class scores $S_b^{\mathcal{C}}$ through the same decoupled expert pipeline during both training and inference.

\subsection{Preliminaries}
\noindent \textbf{Continual Learning.} COD requires a model to learn from a task stream $\mathcal{T}=\{\mathcal{T}_1, \dots, \mathcal{T}_n\}$. At step $t$, the dataset $\mathcal{D}_t=\{\mathcal{X}_i, \mathcal{Y}_i\}$ contains annotations only for the current categories $C_c$, while previously learned categories $C_p$ may still appear without labels. After training on task $t$, the detector is evaluated over all seen categories $C_p\cup C_c$. 

\noindent \textbf{F-ViT Framework.} We build on F-ViT \cite{kuo2022f} framework, a CLIP-based OVD framework that freezes the image and text encoders and trains a standard detection head with text embeddings as classifier weights. While this design provides strong zero-shot detection ability, its trainable head remains shared across categories. Under continual fine-tuning with incomplete annotations, updates for current classes can overwrite the representations previously calibrated for old classes.

\subsection{Cost Volume-Guided Feature Decoupling} \label{sec:cost_volume_decoupling}
The first step of CL-CLIP is to construct category-indexed spatial responses as class-specific priors for continual detection. The key property we exploit is that these responses are derived from the frozen CLIP encoders: since neither the image nor text encoder is updated during continual fine-tuning, the cost volume provides a stable per-category spatial signal that does not degrade as new tasks arrive. Following CAT-Seg \cite{cho2024cat}, originally designed for open-vocabulary semantic segmentation, we compute the cosine similarity between dense CLIP visual tokens and category-specific text embeddings to obtain a Similarity Cost Volume. This cost volume is refined by Swin-based spatial aggregation within each category response and an upsampling decoder, producing category-wise logits maps $\mathbf{Z}_{cv}$:
\begin{equation}
    \mathbf{Z}_{cv}=Cost(\mathcal{F}(I))
\end{equation}
We adapt CAT-Seg for continual detection with two key modifications. First, whereas CAT-Seg uses the cost volume as \emph{final dense predictions} for segmentation, we use it as a \emph{spatial prior} that gates region features for downstream proposal-based detection. Second, we remove CAT-Seg's class aggregation module, which re-introduces cross-category coupling and propagates forgetting signals from unannotated old objects that are treated as background. The resulting category-wise responses are later used by the RoI head to construct class-conditioned RoI representations (detailed in Sec.~\ref{sec:multi_expert_roi}). 

\noindent \textbf{Reducing Cross-Category Spatial Overlap.} Because the response maps are learned without dense mask supervision, different categories may co-activate on the same spatial regions. We introduce a soft orthogonality constraint. Let $\mathbf{P}=\sigma(\mathbf{Z}_{cv})\in\mathbb{R}^{K\times H\times W}$ denote the response probabilities of categories. We flatten and apply $\ell_2$ normalization over the $N=H\times W$ spatial positions to obtain $\tilde{\mathbf{P}}\in\mathbb{R}^{K\times N}$, then compute the Gram matrix $\mathbf{G}=\tilde{\mathbf{P}}\tilde{\mathbf{P}}^\top$ to measure spatial co-activation across class channels, and penalize its off-diagonal entries:
\begin{equation}
    \mathcal{L}_{ortho} = \frac{1}{N_b \cdot K(K-1)}\sum_{n=1}^{N_b}\sum_{i\neq j}(\mathbf{G}_{n,i,j})^2
\end{equation}
% Fig.~\ref{fig:gram_heatmap} visualizes the resulting Gram matrix after the 10+10 VOC sequence (20 classes), showing near-zero off-diagonal values and confirming effective spatial separation. 
We use the mean squared off-diagonal value as a quantitative measure of cross-category spatial entanglement, which we call \textbf{Response Overlap}; Tab.~\ref{tab:response_overlap} analyzes how this metric changes across decoupling variants.

% \begin{figure}[htbp]
%     \centering
%     \includegraphics[width=0.55\textwidth]{figures/gram_matrix_heatmap.pdf}
%     \caption{Gram matrix of spatially $\ell_2$-normalized response maps after the 10+10 PASCAL VOC sequence (20 classes). Off-diagonal mean is $0.0158 \pm 0.0051$, confirming that the orthogonality loss effectively reduces cross-category spatial overlap.}
%     \label{fig:gram_heatmap}
% \end{figure}

\subsection{Spatial Attention-Guided RPN}
In continual detection, the RPN \cite{ren2015faster} may suppress old objects because its objectness targets are defined only by current-task annotations. We use the Similarity Cost Volume as a foreground prior by taking the maximum response over the category dimension to obtain a class-agnostic attention map.

Rather than replacing FPN \cite{lin2017feature} features with this prior, we use a residual gating form:
\begin{equation}
    \mathbf{F}'=\mathbf{F}\times(\beta+\mathbf{A})
\end{equation}
where $\mathbf{F}$ is the FPN feature, $\beta$ is a learnable scalar gate initialized to 1, and $\mathbf{A}$ is the resized attention map. This softly biases the RPN toward regions with high CLIP-text response while preserving the original features. Tab.~\ref{tab:rpn_recall} compares this with a standard RPN, generic spatial attention, and direct cost-volume modulation through proposal recall.

\subsection{Multi-Expert RoI Head} \label{sec:multi_expert_roi}
We keep the standard box regression branch but replace the shared RoI classifier with category-specific experts. For each proposal $b$, RoI-Align extracts a visual feature $b_{vis}\in\mathbb{R}^{d\times d_1\times d_2}$ from the FPN and a per-category spatial gate $b_{seg}\in\mathbb{R}^{K\times d_1\times d_2}$ from the sigmoid-activated cost-volume maps, where each slice $b_{seg}[c]\in\mathbb{R}^{d_1\times d_2}$ is a spatial attention mask for category $c$. Broadcasting $b_{seg}[c]$ across channels yields the class-conditioned feature $b_{dec}[c] = b_{vis} \odot b_{seg}[c]$, so that each of the $K$ slices in $b_{dec}\in\mathbb{R}^{K\times d\times d_1\times d_2}$ retains only the spatially relevant information for its category. Each category $c$ is assigned a convolutional expert $\mathcal{E}_c$ that processes its slice $b_{dec}[c]$. Following Sylph \cite{yin2022sylph}, we replace the shared softmax classifier with per-class sigmoid classifiers, but differ in that Sylph dynamically generates expert weights from few-shot support images over shared RoI features, whereas our experts are directly learned modules that operate on already spatially-gated, class-specific features. Once a category's expert is trained, it is frozen in subsequent tasks, providing architectural protection against forgetting. We compute CLIP-aligned scores:
\begin{equation} \label{detection scores}
    s_b^{c}=\tau\cdot\cos\bigl(\mathcal{E}_c(b_{dec}[c]),\,\mathbf{t}_c\bigr),
\end{equation}
where $\mathbf{t}_c$ is the CLIP text embedding for category $c$ and $\tau$ is the logit scale. At each task $t$, the cost volume and expert pipeline operate over all seen categories $C_p\cup C_c$, so that frozen old-class experts continue to receive their gated inputs. The classification loss $\mathcal{L}_{cls}$ is Binary Cross-Entropy with Logits applied to $\{s_b^c\}_{c\in C_c}$; proposals matching previous categories are excluded from the classification loss rather than treated as negatives. At inference, scoring is applied to all seen categories $C_p\cup C_c$, and the detection score is $S_b^c=\sigma(s_b^c)$. Only current categories' experts are updated; previous categories' experts are frozen. Tab.~\ref{tab:ablation_study_components} examines this design; a t-SNE visualization is provided in Appendix Fig.~\ref{fig:tsne_visualization}.

Because all experts share a fixed logit scale $\tau$ and align to frozen CLIP text embeddings through bounded cosine similarity, old and new experts produce comparable score magnitudes without post-hoc calibration.

Since the decoupled RoI features are still fused with FPN representations, we apply EWC \cite{kirkpatrick2017overcoming} drift regularization ($\mathcal{L}_{drift}$) on the lateral and top-down fusion convolutions of the FPN. The diagonal Fisher information is computed after each task from the total loss gradients, and the drift loss penalizes parameter deviations weighted by importance. Together with the L1 box regression loss ($\mathcal{L}_{reg}$), the full training objective is:
\begin{equation}
    \mathcal{L}=\mathcal{L}_{cls} + \mathcal{L}_{reg} + \mathcal{L}_{ortho} + \lambda\mathcal{L}_{drift}
\end{equation}

%% file: sec/Experiments.tex
\section{Experiments} \label{sec:experiments}
\subsection{Datasets and Evaluation Protocols}
We evaluate CL-CLIP on PASCAL VOC \cite{everingham2010pascal} and MS-COCO \cite{lin2014microsoft}. For VOC, we use the standard two-step splits 10+10, 15+5, and 19+1, where $A+B$ denotes $A$ classes in the first task and $B$ newly introduced classes in the second task. For COCO, we mainly use the 4-task protocol \cite{gupta2022ow} to evaluate long-horizon forgetting, with additional 40+40 and 70+10 two-step results deferred to Appendix \ref{Appendix A}. We report mAP@50 for previously learned classes (mAP@P), current-task classes (mAP@C), and all seen classes (mAP@A).

\subsection{Baseline Construction and Pre-Evaluation} \label{sec:baseline_pre_eval}
Before evaluating CL-CLIP, we first test whether CLIP-based open-vocabulary detectors can naturally handle continual object detection. We build this diagnostic baseline on F-ViT \cite{kuo2022f} by sequentially fine-tuning its detector head on current-task categories while evaluating old categories through mAP@P. This setting directly measures whether the frozen CLIP prior can preserve previous categories after the shared trainable head is updated on new tasks.

\begin{table}[htbp]
    \caption{Performance of various CLIP models on the 4-tasks MS-COCO benchmark}
    \label{tab:baseline_coco_4task}
    \centering
    \resizebox{\textwidth}{!}{
        \begin{tabular}{ll c ccc ccc ccc}
             \toprule
             \multicolumn{1}{c}{\multirow{2}{*}{Model}} & \multicolumn{1}{c}{\multirow{2}{*}{ViT}} & \textbf{Task1} & \multicolumn{3}{c}{\textbf{Task2}} & \multicolumn{3}{c}{\textbf{Task3}} & \multicolumn{3}{c}{\textbf{Task4}} \\
             \cmidrule(lr){3-3} \cmidrule(lr){4-6} \cmidrule(lr){7-9} \cmidrule(lr){10-12}
             & & mAP@C & mAP@C & mAP@P & mAP@A & mAP@C & mAP@P & mAP@A & mAP@C & mAP@P & mAP@A \\
             \midrule
             FineCLIP\cite{jing2024fineclip} & \multirow{4}{*}{ViT-B/16} & 64.3 & 44.5 & \textbf{26.2} & \textbf{35.8} & \textbf{46.1} & \textbf{19.1} & \textbf{28.1} & 38.4 & \textbf{17.4} & \textbf{22.6} \\
             FG-CLIP\cite{xie2025fg} &  & 60.3 & 43.2 & 19.3 & 31.8 & 37.0 & 15.6 & 22.7 & 33.2 & 13.3 & 18.3 \\
             EVA-CLIP\cite{sun2023eva} &  & 61.8 & 43.8 & 5.0 & 25.3 & 39.9 & 10.5 & 20.3 & 35.4 & 8.5 & 15.2 \\
             \midrule
             FineCLIP\cite{jing2024fineclip} & \multirow{4}{*}{ViT-L/14} & 73.7 & 56.2 & \textbf{29.7} & \textbf{43.6} & 57.4 & \textbf{30.8} & \textbf{39.7} & 50.7 & \textbf{27.4} & \textbf{33.2} \\
             FG-CLIP\cite{xie2025fg} &  & 72.6 & 55.4 & 13.3 & 35.4 & 53.4 & 19.0 & 30.5 & 48.3 & 18.1 & 25.6 \\
             EVA-CLIP\cite{sun2023eva} &  & 70.0 & 53.5 & 15.0 & 35.2 & 48.2 & 16.0 & 26.7 & 44.0 & 14.9 & 22.2 \\
             \midrule
             SigLIP2\cite{tschannen2025siglip} & So/14 & \textbf{65.9} & \textbf{48.4} & 7.3 & 28.9 & 45.5 & 9.2 & 21.3 & \textbf{40.4} & 8.2 & 16.3 \\
             \midrule
             SigLIP2\cite{tschannen2025siglip} & So/16 & \textbf{76.5} & \textbf{59.7} & 8.6 & 35.4 & \textbf{58.7} & 13.7 & 28.7 & \textbf{55.8} & 13.0 & 23.7 \\
             \bottomrule
        \end{tabular}
    }
\end{table}

Using this protocol, we benchmark EVA-CLIP\cite{sun2023eva}, SigLIP2\cite{tschannen2025siglip}, FineCLIP\cite{jing2024fineclip}, FG-CLIP\cite{xie2025fg}. Tab.~\ref{tab:baseline_coco_4task} shows that strong CLIP representations do not by themselves solve continual detection. Several variants keep high current-task accuracy but lose old-class performance almost completely, indicating that open-vocabulary generalization and continual retention are not equivalent once the detector is sequentially fine-tuned.

FineCLIP provides the most stable retention and the best mAP@A in most stages, so we use it as the default CLIP backbone. Even this strongest baseline still suffers substantial old-class degradation, motivating an explicitly continual architecture rather than a direct reuse of an OVD detector. Appendix \ref{Appendix A} reports the same pattern on additional two-step COCO and VOC settings.

\subsection{Implementation Details}
We implement CL-CLIP with PyTorch \cite{paszke2019pytorch} and MMDetection \cite{chen1906mmdetection}. Following CAT-Seg \cite{cho2024cat}, we set the main hyperparameters to $d_F=128$, $N_B=2$, and $N_U=2$. The total batch size is 16 for PASCAL VOC and 32 for MS-COCO. We use FineCLIP ViT-B/16 as the default frozen backbone. For the COCO 4-task protocol we follow the class split of OW-DETR \cite{gupta2022ow}; for VOC we use the standard 10+10, 15+5, and 19+1 splits. Multi-seed variance is reported in Appendix~\ref{Appendix E}. Complete optimization settings are deferred to Appendix \ref{Appendix B}.

\subsection{Results and analysis}
\noindent \textbf{Evaluating performance on PASCAL VOC.} Tab. \ref{tab:voc_comparision} summarizes the results on three standard two-step VOC splits. Compared with the FineCLIP/F-ViT baseline, CL-CLIP improves all-class mAP by 7.4, 9.6, and 9.2 points on the 10+10, 15+5, and 19+1 splits, respectively. The gains come from both retention and adaptation: old-class AP increases from 55.8 to 64.8, 65.9 to 78.3, and 68.0 to 77.1, while new-class AP reaches 84.5, 74.5, 84.8. 

Against existing COD methods, CL-CLIP achieves the best all-class mAP on three VOC settings(74.7, 77.4 77.5). Notably, our FineCLIP baseline (67.3/67.8/68.3) underperforms both ResNet-50-based methods with dedicated continual learning mechanisms (ABR 72.0/71.0/70.9, NSGP-RePRE 74.0/73.6/76.0) and IODC (67.6/70.2/72.3), which shares the same CLIP ViT-B/16 backbone but uses different pretraining. Although these comparisons involve different architectures and strategies, they collectively suggest that stronger pretraining alone does not confer continual retention. The gains of CL-CLIP therefore stem from the category decoupling design rather than backbone capacity. Compared to IODC, CL-CLIP improves all-class mAP from 67.6/70.2/72.3 to 74.7/77.4/77.5. It also compares favorably with parameter-isolation methods such as TARL \cite{zhang2024learning} and ROSETTA \cite{yang2022continual}, achieving higher new-class AP while keeping competitive old-class AP, and obtains a better stability-plasticity trade-off than NSGP-RePRE \cite{wu2025demystifying}. All comparison numbers except GCD \cite{wang2025gcd} are taken from original publications; GCD is reproduced with released code as it does not report VOC results.

\begin{table}[htbp]
    \centering
    \caption{mAP@0.5 results on single continual step on PASCAL VOC}
    \label{tab:voc_comparision}
    \footnotesize
    \setlength{\tabcolsep}{4pt}
    \begin{tabular}{ll  ccc  ccc  ccc}
        \toprule
        \multirow{2}{*}{\textbf{Method}} & \multirow{2}{*}{\textbf{Backbone}} & \multicolumn{3}{c}{\textbf{10+10}} & \multicolumn{3}{c}{\textbf{15+5}} & \multicolumn{3}{c}{\textbf{19+1}} \\
        \cmidrule(lr){3-5} \cmidrule(lr){6-8} \cmidrule(lr){9-11}
        & & \textbf{1-10} & \textbf{11-20} & \textbf{1-20} & \textbf{1-15} & \textbf{16-20} & \textbf{1-20} & \textbf{1-19} & \textbf{20} & \textbf{1-20} \\
        \midrule
        Baseline & CLIP ViT-B/16 & 55.8 & 78.8 & 67.3 & 65.9 & 73.4 & 67.8 & 68.0 & 74.3 & 68.3 \\
        \midrule
        OW-DETR\cite{gupta2022ow} & ResNet-50 & 63.5 & 67.9 & 65.7 & 72.2 & 59.8 & 69.4 & 71.1 & 62 & 70.2 \\
        ABR\cite{liu2023augmented} & ResNet-50 & 71.2 & 72.8 & 72.0 & 73.0 & 65.1 & 71.0 & 71.0 & 69.7 & 70.9 \\
        MMA\cite{cermelli2022modeling} & ResNet-50 & 69.3 & 63.9 & 66.6 & 73.0 & 60.5 & 69.9 & 71.1 & 63.4 & 70.7 \\
        NSGP-RePRE\cite{wu2025demystifying} & ResNet-50 & \textbf{75.3} & 72.7 & 74 & 77.5 & 61.8 & 73.6 & 76.3 & 69 & 76 \\
        ROSETTA\cite{yang2022continual} & ResNet-50 & 67.6 & 66.0 & 66.8 & 71.7 & 61.6 & 69.2 & 69.7 & 68.3 & 69.6 \\
        GCD\cite{wang2025gcd} & Swin-T & 67.3 & 71.4 & 69.4 & 69.5 & 58.8 & 66.8 & 69.8 & 74.6 & 70.1 \\
        \midrule
        TARL\cite{zhang2024learning} & Swin-T & 71.2 & 70.0 & 70.6 & 73.6 & 60.2 & 70.3 & 73.2 & 66.5 & 72.9 \\
        IODC\cite{huang2023incremental} & CLIP ViT-B/16 & 66.1 & 69.2 & 67.6 & 73.1 & 61.2 & 70.2 & 72.8 & 62.5 & 72.3 \\
        \midrule
        CL-CLIP(Ours) & CLIP ViT-B/16 & 64.8 & \textbf{84.5} & \textbf{74.7} & \textbf{78.3} & \textbf{74.5} & \textbf{77.4} & \textbf{77.1} & \textbf{84.8} & \textbf{77.5} \\
        \bottomrule
    \end{tabular}
\end{table}

\begin{figure}
    \centering
    \includegraphics[width=\textwidth]{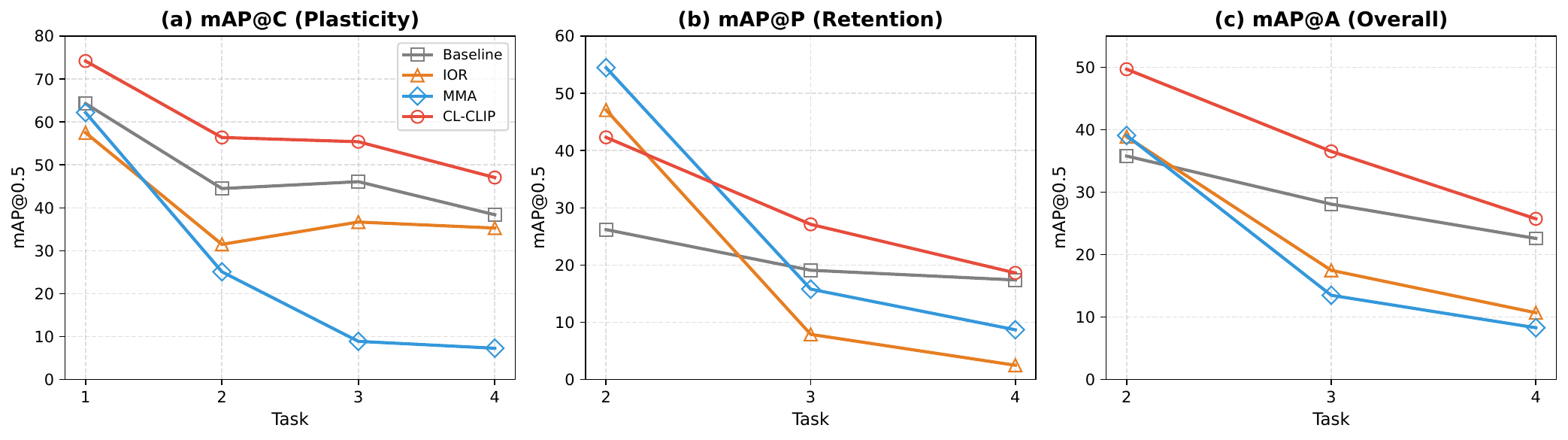}
    \caption{Performance on the 4-task MS-COCO benchmark (mAP@0.5). CL-CLIP maintains stronger plasticity and overall accuracy across the full task sequence compared with reproduced baselines.}
    \label{fig:coco_curve}
\end{figure}

\noindent \textbf{Evaluating performance on MS-COCO.} The 4-task MS-COCO protocol has not been widely adopted by COD methods, most of which only report two-step COCO results. To examine long-horizon forgetting, we reproduce IOR \cite{an2025ior} and MMA \cite{cermelli2022modeling} under this protocol. Fig.~\ref{fig:coco_curve} plots mAP across the four sequential tasks.

CL-CLIP improves the FineCLIP/F-ViT baseline throughout the stream, raising mAP@A from 35.8 to 49.7 (Task2), 28.1 to 36.5 (Task3), and 22.6 to 25.7 (Task4), while maintaining a more balanced stability-plasticity trade-off than IOR and MMA across the full sequence. Appendix~\ref{Appendix E} confirms the same trend under the stricter mAP@[.5:.95] metric.

% \begin{table}[htbp]
%     \centering
%     \caption{mAP@0.5 results on multiple continual steps on 4-task MS-COCO}
%     \label{tab:coco_comparision}
%     \resizebox{\textwidth}{!}{
%     \begin{tabular}{l c ccc ccc ccc}
%          \toprule
%          \multicolumn{1}{c}{\multirow{2}{*}{\textbf{Methods}}} & \textbf{Task1} & \multicolumn{3}{c}{\textbf{Task2}} & \multicolumn{3}{c}{\textbf{Task3}} & \multicolumn{3}{c}{\textbf{Task4}} \\
%          \cmidrule(lr){2-2} \cmidrule(lr){3-5} \cmidrule(lr){6-8} \cmidrule(lr){9-11}
%          & mAP@C & mAP@C & mAP@P & mAP@A & mAP@C & mAP@P & mAP@A & mAP@C & mAP@P & mAP@A \\
%          \midrule
%          Baseline & 64.3 & 44.5 & 26.2 & 35.8 & 46.1 & 19.1 & 28.1 & 38.4 & 17.4 & 22.6 \\
%          \midrule
%          IOR\cite{an2025ior} & 57.5 & 31.5 & \textbf{47.1} & 38.9 & 36.7 & 7.9 & 17.5 & 35.3 & 2.5 & 10.7 \\
%          MMA\cite{cermelli2022modeling} & 62.2 & 54.5 & 25.1 & 39.1 & 8.9 & 15.8 & 13.5 & 7.3 & 8.7 & 8.3 \\
%          CL-CLIP(Ours) & \textbf{74.2} & \textbf{56.4} & 42.3 & \textbf{49.7} & \textbf{55.4} & 27.1 & \textbf{36.5} & \textbf{47.1} & 18.6 & 25.7 \\
%          \bottomrule 
%     \end{tabular}
%     }
% \end{table}

\subsection{Ablation study} \label{sec:ablation}
\begin{figure}
    \centering
    \includegraphics[width=\linewidth]{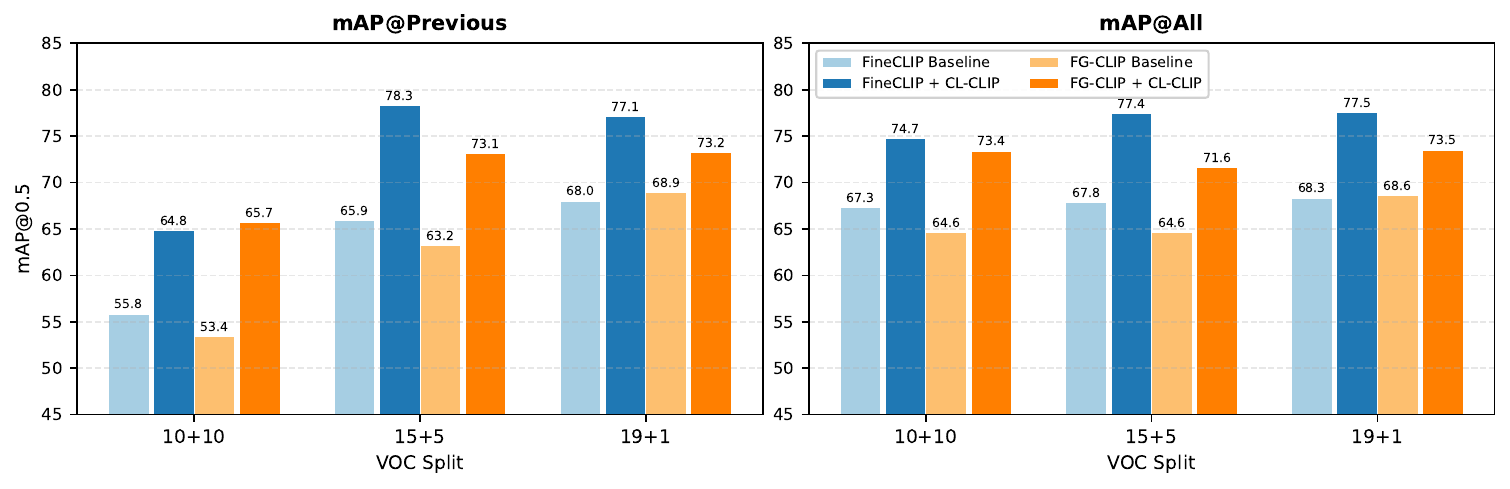}
    \caption{Generalization of CL-CLIP across CLIP backbones on PASCAL VOC (mAP@0.5). Left: previous-task classes; Right: all classes. CL-CLIP consistently improves retention and overall accuracy on both FineCLIP and FG-CLIP.}
    \label{fig:clip_backbone}
\end{figure}

\noindent \textbf{Generalization across CLIP backbones.} To verify that CL-CLIP is not tied to a specific CLIP variant, we evaluate it with FG-CLIP \cite{xie2025fg} ViT-B/16 in addition to the default FineCLIP backbone. Fig.~\ref{fig:clip_backbone} shows that CL-CLIP consistently improves both backbones across all three VOC splits, confirming that the category decoupling design generalizes across different CLIP representations.

\noindent \textbf{Core mechanism ablation.} Tab.~\ref{tab:decoupling_drift} isolates the two core mechanisms: cost-volume-guided category decoupling and EWC-style drift regularization. Removing decoupling (i.e., falling back to the shared F-ViT head while keeping drift regularization) sharply reduces both plasticity and retention, confirming that drift regularization alone cannot substitute for class-specific pathways. Removing drift regularization preserves plasticity but causes severe retention collapse (mAP@P drops from 42.3 to 14.9 on Task2). Only combining both achieves a balanced trade-off.

\begin{table}[htbp]
    \centering
    \caption{Core mechanism ablation on the 4-task MS-COCO benchmark: cost-volume decoupling and drift regularization}
    \label{tab:decoupling_drift}
    \resizebox{\textwidth}{!}{
    \begin{tabular}{cc ccc ccc ccc}
         \toprule
         \multicolumn{2}{c}{\textbf{Components}} & \multicolumn{3}{c}{\textbf{Task2}} & \multicolumn{3}{c}{\textbf{Task3}} & \multicolumn{3}{c}{\textbf{Task4}} \\
         \cmidrule(lr){1-2} \cmidrule(lr){3-5} \cmidrule(lr){6-8} \cmidrule(lr){9-11}
         Decoupling & Drift Reg. & mAP@C & mAP@P & mAP@A & mAP@C & mAP@P & mAP@A & mAP@C & mAP@P & mAP@A \\
         \midrule
         \checkmark & \checkmark & 56.4 & \textbf{42.3} & \textbf{49.7} & 55.4 & \textbf{27.1} & \textbf{36.5} & 47.1 & \textbf{18.6} & \textbf{25.7} \\
          & \checkmark & 50.4 & 20.1 & 36.0 & 49.6 & 20.4 & 30.2 & 43.8 & 15.8 & 22.8 \\
         \checkmark &  & \textbf{56.6} & 14.9 & 36.8 & \textbf{55.9} & 10.6 & 25.7 & \textbf{50.2} & 3.9 & 15.4 \\
         \bottomrule
    \end{tabular}
    }
\end{table}

\noindent \textbf{Structural component ablation.} We evaluate architectural components on the 4-task MS-COCO benchmark (Tab. \ref{tab:ablation_study_components}). Adding class aggregation back slightly improves current-class AP but severely damages old-class retention. On Task4, mAP@P falls from 18.6 to 5.4 and mAP@A drops from 25.7 to 16.5. This confirms that cross-class aggregation is harmful under missing-label supervision, as it encourages competition between annotated new classes and unannotated old classes.

Removing convolutional experts and falling back to the F-ViT head (with FPN regularization retained) drops mAP@A to 36.0/30.2 on Task2/Task3 vs 49.7/36.5 for CL-CLIP, confirming that FPN regularization alone is insufficient without class-specific pathways. The spatial-attention-guided RPN has a more modest effect; we view it as complementary to class aggregation removal and the Multi-Expert branch.

\begin{table}[htbp]
    \caption{Ablation study of proposed structure components on the 4-task MS-COCO benchmark}
    \label{tab:ablation_study_components}
    \centering
    \resizebox{\textwidth}{!}{
    \begin{tabular}{c c c c ccc ccc ccc}
         \toprule
         \multicolumn{3}{c}{\textbf{Modules}} & \textbf{Task1} & \multicolumn{3}{c}{\textbf{Task2}} & \multicolumn{3}{c}{\textbf{Task3}} & \multicolumn{3}{c}{\textbf{Task4}} \\
         \cmidrule(lr){1-3} \cmidrule(lr){4-4} \cmidrule(lr){5-7} \cmidrule(lr){8-10} \cmidrule(lr){11-13}
         \makecell{class \\ aggregation} & \makecell{spatial \\ -RPN} & \makecell{conv \\ expert} & mAP@C & mAP@C & mAP@P & mAP@A & mAP@C & mAP@P & mAP@A & mAP@C & mAP@P & mAP@A \\
         \midrule
         & \checkmark & \checkmark & \textbf{74.2} & 56.4 & \textbf{42.3} & \textbf{49.7} & 55.4 & \textbf{27.1} & \textbf{36.5} & 47.1 & 18.6 & 25.7 \\
         \checkmark & \checkmark & \checkmark & \textbf{74.2} & \textbf{56.5} & 4.1 & 31.6 & \textbf{55.5} & 10.2 & 25.3 & \textbf{49.8} & 5.4 & 16.5 \\
         & \checkmark &  & 69.7 & 50.4 & 20.1 & 36.0 & 49.6 & 20.4 & 30.2 & 43.8 & 15.8 & 22.8 \\
         & & \checkmark & 74.0 & 55.9 & 42.0 & 49.3 & 55.0 & 26.8 & 36.2 & 48.4 & \textbf{19.2} & \textbf{26.5} \\
         \bottomrule
    \end{tabular}
    }
\end{table}

\noindent \textbf{Category response overlap.} To examine cross-category spatial entanglement,  Tab.~\ref{tab:response_overlap} reports the Response Overlap metric. Without the orthogonality loss, overlap reaches 0.9945 because the trainable aggregation modules, optimized solely by the detection loss, converge toward a category-agnostic foreground pattern that erases the per-class separation present in the raw CLIP similarities. The orthogonality loss counteracts this by penalizing spatial co-activation, reducing overlap to 0.0017 and improving average old-class AP from 23.5 to 29.3. The class aggregation variant shows moderate overlap (0.0112) but much lower retention (Avg.\ mAP@P 6.6), confirming that cross-category coupling harms old-class preservation through a different mechanism.

\begin{table}[htbp]
    \centering
    \caption{Response Overlap and average detection performance over Task2--Task4 of 4-task MS-COCO. Response Overlap is the mean squared off-diagonal Gram value of sigmoid response maps after spatial $\ell_2$ normalization (lower is better).}
    \label{tab:response_overlap}
    \resizebox{\textwidth}{!}{
    \begin{tabular}{l cccc}
         \toprule
         \textbf{Variant} & \textbf{Avg. mAP@P} & \textbf{Avg. mAP@C} & \textbf{Avg. mAP@A} & \textbf{Response Overlap} $\downarrow$ \\
         \midrule
         w/ class aggregation & 6.6 & 53.9 & 24.5 & 0.0112 \\
         w/o ortho loss & 23.5 & 53.7 & 34.4 & 0.9945 \\
         Full CL-CLIP & 29.3 & 53.0 & 37.3 & 0.0017 \\
         \bottomrule
    \end{tabular}
    }
\end{table}

\noindent \textbf{Proposal recall of Spatial-RPN.} Tab.~\ref{tab:rpn_recall} compares standard RPN, CBAM-style spatial attention \cite{woo2018cbam}, direct cost-volume modulation, and our residual-gated cost-volume prior on VOC 10+10. Direct cost-volume modulation obtains the highest proposal recall, suggesting that the cost-volume prior improves object coverage, but its detection accuracy is lower than the residual-gated design. Our residual-gated prior achieves the best old-class and all-class mAP while keeping competitive recall, supporting it as a minor but complementary refinement to the main decoupling mechanism.

\begin{table}[htbp]
    \centering
    \caption{Proposal recall analysis for Spatial-RPN on the PASCAL VOC 10+10 setting. R@100 is computed at IoU 0.5 using top-100 RPN proposals per image.}
    \label{tab:rpn_recall}
    \resizebox{\textwidth}{!}{
    \begin{tabular}{l ccccc}
         \toprule
         \textbf{RPN Variant} & \textbf{mAP@P} & \textbf{mAP@A} & \textbf{R@100-P} & \textbf{R@100-C} & \textbf{R@100-A} \\
         \midrule
         Standard RPN & 62.7 & 74.1 & 0.320 & \textbf{0.571} & 0.475 \\
         CBAM-style spatial attention & 63.3 & 74.4 & 0.327 & 0.570 & 0.478 \\
         Direct cost-volume modulation & 63.4 & 74.2 & \textbf{0.397} & 0.562 & \textbf{0.499} \\
         Residual-gated cost-volume prior & \textbf{64.8} & \textbf{74.7} & 0.356 & 0.567 & 0.487 \\
         \bottomrule
    \end{tabular}
    }
\end{table}

\noindent \textbf{Auxiliary loss ablation.} Tab.~\ref{tab:ablation_study_loss} (Appendix~\ref{Appendix F}) studies the auxiliary losses. Drift regularization is critical for FPN stability; combined with the orthogonality loss it yields the best overall accuracy, as the two losses complement each other in stabilizing shared features and maintaining separable cost-volume responses.

\noindent \textbf{Implementation choices.} Appendix \ref{Appendix C} compares alternative regularization strategies and expert structures, supporting our use of EWC and convolutional experts.

\section{Conclusion}
We studied continual object detection from the perspective of CLIP-based open-vocabulary detection. Our baseline analysis shows that the zero-shot advantage of CLIP-based detectors does not automatically translate into continual learning ability. CL-CLIP addresses this by using a CLIP-text similarity cost volume as a zero-shot spatial prior to build class-specific detection pathways, reducing cross-class interference. Experiments on PASCAL VOC and MS-COCO confirm substantially improvements over the F-ViT baseline and competitive performance compared with existing COD methods, suggesting that category decoupling is a promising direction for CLIP-based continual detection. 

\paragraph{Limitations.}
We report multi-seed variance on one representative split (VOC 10+10, Tab.~\ref{tab:multi_seed}); extending to all settings remains costly. Evaluation is limited to PASCAL VOC and MS-COCO; generalization to larger-scale or longer-horizon scenarios remains to be validated. The Multi-Expert RoI head introduces one expert per category, so inference FLOPs grow linearly with the number of seen classes (Appendix~\ref{Appendix E}).

%% file: sec/appendix.tex
\appendix

\section{Baseline Results on Two-Stage Benchmarks}\label{Appendix A}
Tab. \ref{tab:baseline_coco_2task} and Tab. \ref{tab:baseline_voc_2task} present additional baseline pre-evaluation results across CLIP variants on two-step MS-COCO and PASCAL VOC settings, respectively. These results complement the 4-task MS-COCO analysis in Sec. \ref{sec:baseline_pre_eval} and examine whether the same forgetting pattern holds under shorter continual sequences.

The results are consistent with the main text. FineCLIP generally provides stronger old-class retention and all-class accuracy than the other evaluated CLIP variants, which supports its use as the default backbone for CL-CLIP. However, all variants still show clear degradation on previously learned classes after sequential fine-tuning. This confirms that stronger open-vocabulary representations alone are insufficient for continual object detection.

\begin{table}[htbp]
    \caption{Performance of various CLIP models on the two-stage MS-COCO benchmark}
    \label{tab:baseline_coco_2task}
    \centering
    \begin{tabular}{c  l l c c c c}
         \toprule
         \multicolumn{1}{c}{\multirow{2}{*}{Setting}} & 
         \multicolumn{1}{c}{\multirow{2}{*}{Model}} & \multicolumn{1}{c}{\multirow{2}{*}{ViT}} & \textbf{Task1} & \multicolumn{3}{c}{\textbf{Task2}}\\
         \cmidrule(lr){4-4} \cmidrule(lr){5-7}
         & & & mAP@C & mAP@C & mAP@P & mAP@A\\
         \midrule
         \multirow{10}{*}{40+40} & FineCLIP\cite{jing2024fineclip} & \multirow{4}{*}{ViT-B/16} & 57.2 & 46.9 & 17.1 & 32.0  \\
         & FG-CLIP\cite{xie2025fg} &  & 49.4 & 35.4 & 16.8 & 26.1 \\
         & EVA-CLIP\cite{sun2023eva} &  & 52.0 & 36.6 & 11.5 & 24.1 \\
         \cmidrule{2-7}
         & FineCLIP\cite{jing2024fineclip} & \multirow{4}{*}{ViT-L/14} & 65.5 & 56.3 & 34.2 & 45.3 \\
         & FG-CLIP\cite{xie2025fg} &  & 62.2 & 50.7 & 19.5 & 35.1 \\
         & EVA-CLIP\cite{sun2023eva} &  & 60.9 & 46.1 & 15.7 & 30.9 \\
         \cmidrule{2-7}
         & SigLIP2\cite{tschannen2025siglip} & So/14 & 55.4 & 42.9 & 12.3 & 27.6 \\
         \cmidrule{2-7}
         & SigLIP2\cite{tschannen2025siglip} & So/16 & 66.1 & 56.5 & 14.8 & 35.7 \\
         \midrule 
         \midrule
         \multirow{10}{*}{70+10} & FineCLIP\cite{jing2024fineclip} & \multirow{4}{*}{ViT-B/16} & 52.7 & 34.8 & 24.1 & 25.5  \\
         & FG-CLIP\cite{xie2025fg} &  & 43.7 & 25.0 & 21.2 & 21.7 \\
         & EVA-CLIP\cite{sun2023eva} &  & 46.1 & 25.9 & 17.2 & 18.3 \\
         \cmidrule{2-7}
         & FineCLIP\cite{jing2024fineclip} & \multirow{4}{*}{ViT-L/14} & 61.1 & 46.6 & 36.9 & 38.1 \\
         & FG-CLIP\cite{xie2025fg} &  & 57.8 & 40.6 & 21.2 & 21.7 \\
         & EVA-CLIP\cite{sun2023eva} &  & 55.0 & 37.7 & 28.4 & 29.5 \\
         \cmidrule{2-7}
         & SigLIP2\cite{tschannen2025siglip} & So/14 & 51.1 & 31.7 & 20.3 & 21.8 \\
         \cmidrule{2-7}
         & SigLIP2\cite{tschannen2025siglip} & So/16 & 62.6 & 48.6 & 26.1 & 28.9 \\
         \bottomrule
    \end{tabular}
\end{table}

\begin{table}[htbp]
    \caption{Performance of various CLIP models on the two-stage PASCAL VOC benchmark}
    \label{tab:baseline_voc_2task}
    \centering
    \resizebox{\textwidth}{!}{
        \begin{tabular}{l l c ccc c ccc c ccc}
            \toprule
            \multicolumn{1}{c}{\multirow{3}{*}{Model}} & \multicolumn{1}{c}{\multirow{3}{*}{ViT}} & \multicolumn{4}{c}{\textbf{10+10}} & \multicolumn{4}{c}{\textbf{15+5}} & \multicolumn{4}{c}{\textbf{19+1}} \\
            \cmidrule(lr){3-6} \cmidrule(lr){7-10} \cmidrule(lr){11-14}
            & & \textbf{Task1} & \multicolumn{3}{c}{\textbf{Task2}} & \textbf{Task1} & \multicolumn{3}{c}{\textbf{Task2}} & \textbf{Task1} & \multicolumn{3}{c}{\textbf{Task2}} \\
            \cmidrule(lr){3-3} \cmidrule(lr){4-6} \cmidrule(lr){7-7} \cmidrule(lr){8-10} \cmidrule(lr){11-11} \cmidrule(lr){12-14}
            & & mAP@C & mAP@C & mAP@P & mAP@A & mAP@C & mAP@C & mAP@P & mAP@A & mAP@C & mAP@C & mAP@P & mAP@A \\
            \midrule
            FineCLIP\cite{jing2024fineclip} & \multirow{4}{*}{ViT-B/16} & 80.2 & 78.8 & 55.8 & 67.3 & 81.4 & 73.4 & 65.9 & 67.8 & 79.5 & 74.3 & 68.0 & 68.3 \\
            FG-CLIP\cite{xie2025fg} & & 76.4 & 75.7 & 53.4 & 64.6 & 77.0 & 69.0 & 63.2 & 64.6 & 76.4 & 62.1 & 68.9 & 68.6 \\
            EVA-CLIP\cite{sun2023eva} & & 76.7 & 78.7 & 37.2 & 58.0 & 79.5 & 73.6 & 55.4 & 59.9 & 78.5 & 74.5 & 64.0 & 64.6 \\
            \midrule
            FineCLIP\cite{jing2024fineclip} & \multirow{4}{*}{ViT-L/14} & 84.2 & 82.8 & 62.0 & 72.4 & 84.3 & 78.6 & 74.2 & 75.3 & 83.4 & 80.9 & 73.3 & 73.6 \\
            FG-CLIP\cite{xie2025fg} & & 84.1 & 82.9 & 53.8 & 68.4 & 84.0 & 76.6 & 73.1 & 74.0 & 83.5 & 74.6 & 76.9 & 76.8 \\
            EVA-CLIP\cite{sun2023eva} & & 83.0 & 83.5 & 39.1 & 61.3 & 84.6 & 80.2 & 66.1 & 69.6 & 83.4 & 76.9 & 73.8 & 74.0 \\
            \midrule
            SigLIP2\cite{tschannen2025siglip} & So/14 & 81.7 & 81.9 & 48.2 & 65.1 & 82.1 & 76.5 & 60.9 & 64.8 & 81.3 & 78.3 & 67.8 & 68.3 \\
            \midrule
            SigLIP2\cite{tschannen2025siglip} & So/16 & 85.9 & 84.8 & 47.5 & 66.1 & 86.5 & 82.2 & 66.3 & 70.3 & 84.6 & 82.5 & 70.8 & 71.3 \\
            \bottomrule
        \end{tabular}
    }
\end{table}

\section{Implementation Details}\label{Appendix B}
All experiments are conducted on 4$\times$ NVIDIA RTX 4090 GPUs. The experimental hyperparameters of our proposed method are shown in Tab.~\ref{tab:implementation_details}. Notably, with the exception of the aggregation module in CAT-Seg, Group Normalization \cite{wu2018group} with the number of groups set to 32 is universally adopted throughout our detector head.

\begin{table}[htbp]
    \caption{Implementation details of CL-CLIP}
    \label{tab:implementation_details}
    \centering
    \begin{tabular}{lc}
         \toprule
         \textbf{Parameter} & \textbf{Value} \\
         \midrule
         % $\alpha$ & 0.1 \\
         % $\beta$ & 0.8 \\
         $\tau$ & 50.0 \\
         % T & 75.0 \\
         Optimizer & AdamW \\ 
         Learning Rate & $4\times 10^{-4}$ \\
         Weight Decay & 0.1 \\
         Learning Rate Schedule & Cosine Annealing \\
         Training Epochs & 20 \\
         Warmup Epochs & 4 \\
         $\lambda$ (drift reg. weight) & 1000 \\
         Norm Cfg & GroupNorm \\
         Backbone & FineCLIP ViT-B/16 \\
         Batch Size (VOC / COCO) & 16 / 32 \\
         Input Resolution & $384\times 384$ \\
         Random Seed & 0 \\
         \bottomrule
    \end{tabular}
\end{table}

\paragraph{Task class splits.}
For the COCO 4-task protocol we follow the class ordering of OW-DETR \cite{gupta2022ow}, partitioned into four groups of 19+21+20+20 categories:
\textbf{Task\,1} (1--19): airplane, bicycle, bird, boat, bus, car, cat, cow, dog, horse, motorcycle, sheep, train, elephant, bear, zebra, giraffe, truck, person;
\textbf{Task\,2} (20--40): traffic light, fire hydrant, stop sign, parking meter, bench, chair, dining table, potted plant, backpack, umbrella, handbag, tie, suitcase, microwave, oven, toaster, sink, refrigerator, bed, toilet, couch;
\textbf{Task\,3} (41--60): frisbee, skis, snowboard, sports ball, kite, baseball bat, baseball glove, skateboard, surfboard, tennis racket, banana, apple, sandwich, orange, broccoli, carrot, hot dog, pizza, donut, cake;
\textbf{Task\,4} (61--80): laptop, mouse, remote, keyboard, cell phone, book, clock, vase, scissors, teddy bear, hair drier, toothbrush, wine glass, cup, fork, knife, spoon, bowl, tv, bottle.

For PASCAL VOC, the 20 categories follow the alphabetical order: aeroplane, bicycle, bird, boat, bottle, bus, car, cat, chair, cow, dining table, dog, horse, motorbike, person, potted plant, sheep, sofa, train, tv monitor. The three splits are 10+10, 15+5, and 19+1, partitioned sequentially in this order.

\section{Additional Ablations on Regularization and Expert Design}\label{Appendix C}
Tab. \ref{tab:ablation_study_regu_moe} provides additional analysis on two implementation choices in CL-CLIP. We compare different regularization strategies for stabilizing the FPN and evaluate alternative expert structures in the Multi-Expert RoI head.

For FPN stabilization, we compare EWC \cite{kirkpatrick2017overcoming} with gradient-space constraint methods including GDA \cite{luo2025gradient} and NSGP \cite{wu2025demystifying}. EWC penalizes updates to parameters that are important for previous tasks, while GDA and NSGP modify the optimization trajectory by constraining gradients during backpropagation. The upper part of Tab. \ref{tab:ablation_study_regu_moe} shows that EWC is more suitable for our FPN stabilizer. GDA and NSGP preserve high current-class AP, but their old-class AP drops sharply in later tasks, which leads to much lower overall accuracy.

We also compare two expert structures in the Multi-Expert RoI head. The convolutional expert processes the RoI feature map with a compact convolutional residual network, while the residual MLP expert first applies average pooling and then uses a 3-layer residual MLP. The lower part of Tab. \ref{tab:ablation_study_regu_moe} shows a clear advantage for convolutional experts. This suggests that preserving local spatial structure inside each RoI is important for class-specific detection.

\begin{table}[htbp]
    \centering
    \caption{Ablation studies on alternative architectural design choices}
    \label{tab:ablation_study_regu_moe}
    \resizebox{\textwidth}{!}{
    \begin{tabular}{c l c ccc ccc ccc}
         \toprule
         \multirow{2}{*}{\textbf{Component}} & \multirow{2}{*}{\textbf{Methods}} & \textbf{Task1} & \multicolumn{3}{c}{\textbf{Task2}} & \multicolumn{3}{c}{\textbf{Task3}} & \multicolumn{3}{c}{\textbf{Task4}} \\
         \cmidrule(lr){3-3} \cmidrule(lr){4-6} \cmidrule(lr){7-9} \cmidrule(lr){10-12}
         & & mAP@C & mAP@C & mAP@P & mAP@A & mAP@C & mAP@P & mAP@A & mAP@C & mAP@P & mAP@A \\
         \midrule
         
         \multirow{3}{*}{Regularization} 
         & EWC \cite{kirkpatrick2017overcoming} & 74.2 & 56.4 & 42.3 & 49.7 & 55.4 & 27.1 & 36.5 & 47.1 & 18.6 & 25.7 \\
         & GDA \cite{luo2025gradient} & 74.2 & 55.5 & 1.2 & 29.7 & 55.2 & 2.1 & 19.8 & 49.5 & 2.6 & 14.4 \\
         & NSGP \cite{wu2025demystifying} & 74.2 & 57.2 & 9.0 & 34.3 & 56.3 & 5.2 & 22.2 & 49.8 & 1.7 & 13.8 \\
         \midrule
         \midrule
         \multirow{2}{*}{Expert Type} 
         & Conv Expert & 74.2 & 56.4 & 42.3 & 49.7 & 55.4 & 27.1 & 36.5 & 47.1 & 18.6 & 25.7 \\
         & Residual MLP Expert & 70.2 & 51.1 & 4.1 & 28.8 & 49.8 & 6.3 & 20.8 & 44.0 & 3.2 & 13.4 \\
         \bottomrule 
    \end{tabular}
    }
\end{table}

\section{Feature-Space Visualization}\label{Appendix D}

\begin{figure}[htbp]
    \centering
    \includegraphics[width=\textwidth]{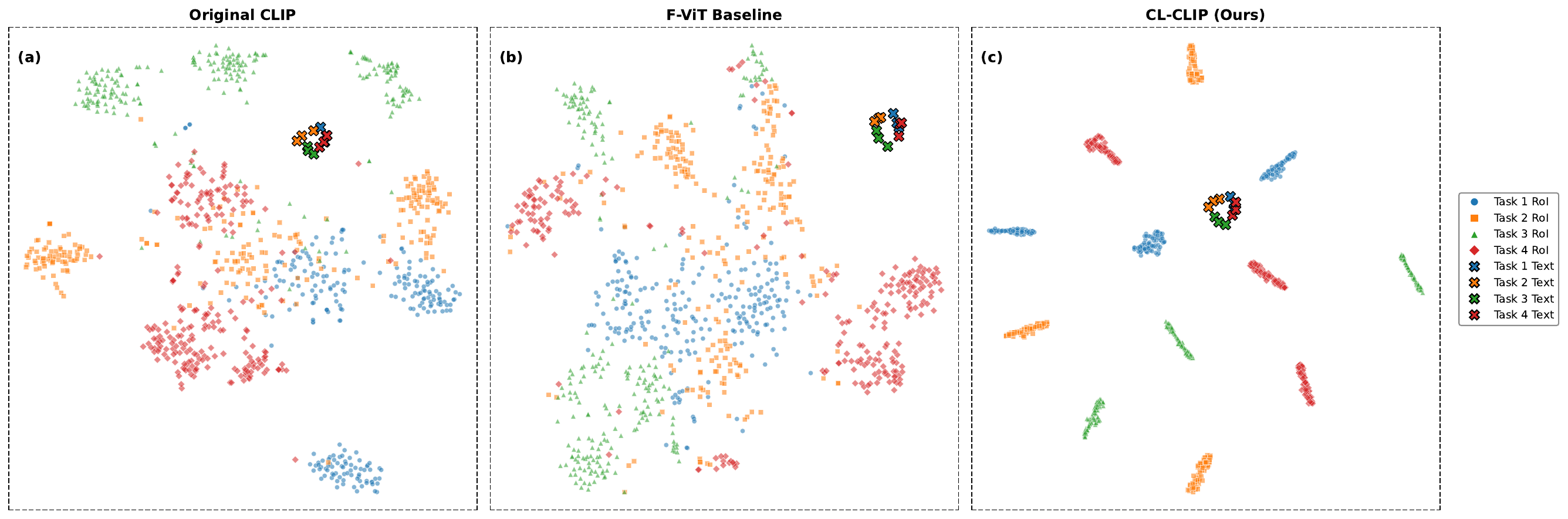}
    \caption{t-SNE visualization of RoI features and CLIP text embeddings after the 4-task MS-COCO sequence. Compared with original FineCLIP and the F-ViT baseline, CL-CLIP forms more compact and better separated RoI feature clusters, suggesting a more structured RoI-text feature space.}
    \label{fig:tsne_visualization}
\end{figure}

To qualitatively examine how the representation space changes after continual fine-tuning, we visualize the RoI features and text embeddings with t-SNE in Fig.~\ref{fig:tsne_visualization}. Original FineCLIP already shows moderate semantic organization, suggesting that the frozen CLIP representation provides a useful prior but insufficient separation for continual detection. After F-ViT fine-tuning, RoI features become more dispersed and entangled, and text embeddings move away from most visual clusters. In contrast, CL-CLIP forms more compact class-wise RoI clusters with clearer separation, consistent with the intended effect of per-class experts and frozen old-class experts.

\section{Auxiliary Loss Ablation}\label{Appendix F}
Tab.~\ref{tab:ablation_study_loss} reports the full loss component ablation on the 4-task MS-COCO benchmark.

\begin{table}[htbp]
    \centering
    \caption{Ablation study of loss components on the 4-task MS-COCO benchmark}
    \label{tab:ablation_study_loss}
    \resizebox{\textwidth}{!}{
    \begin{tabular}{l c ccc ccc ccc}
         \toprule
         \multicolumn{1}{c}{\multirow{2}{*}{\textbf{Losses}}} & \textbf{Task1} & \multicolumn{3}{c}{\textbf{Task2}} & \multicolumn{3}{c}{\textbf{Task3}} & \multicolumn{3}{c}{\textbf{Task4}} \\
         \cmidrule(lr){2-2} \cmidrule(lr){3-5} \cmidrule(lr){6-8} \cmidrule(lr){9-11}
         & mAP@C & mAP@C & mAP@P & mAP@A & mAP@C & mAP@P & mAP@A & mAP@C & mAP@P & mAP@A \\
         \midrule
         cls+reg & \textbf{74.6} & \textbf{57.6} & 5.9 & 33.0 & \textbf{56.4} & 4.7 & 21.9 & 49.9 & 4.4 & 15.8 \\
         cls+reg+drift & \textbf{74.6} & 56.1 & 29.1 & 43.3 & 55.5 & 23.2 & 34.0 & 49.5 & 18.2 & \textbf{26.0} \\
         cls+reg+ortho & 74.2 & 56.6 & 14.9 & 36.8 & 55.9 & 10.6 & 25.7 & \textbf{50.2} & 3.9 & 15.4 \\
         cls+reg+ortho+drift & 74.2 & 56.4 & \textbf{42.3} & \textbf{49.7} & 55.4 & \textbf{27.1} & \textbf{36.5} & 47.1 & \textbf{18.6} & 25.7 \\
         \bottomrule 
    \end{tabular}
    }
\end{table}

\section{Additional Experiments} \label{Appendix E}
\paragraph{Stricter evaluation metric.}
To verify that the improvements are not limited to mAP@0.5, we report the stricter COCO-standard mAP@[.5:.95] under the same 4-task protocol. Tab.~\ref{tab:coco_strict} shows that CL-CLIP consistently outperforms both IOR and MMA on overall accuracy (mAP@A) at every stage, confirming that the gains transfer to the more demanding localization threshold.

\begin{table}[htbp]
    \centering
    \caption{4-task MS-COCO results under the strict mAP@[.5:.95] metric}
    \label{tab:coco_strict}
    \resizebox{\textwidth}{!}{
    \begin{tabular}{l c ccc ccc ccc}
         \toprule
         \multirow{2}{*}{\textbf{Method}} & \textbf{Task1} & \multicolumn{3}{c}{\textbf{Task2}} & \multicolumn{3}{c}{\textbf{Task3}} & \multicolumn{3}{c}{\textbf{Task4}} \\
         \cmidrule(lr){2-2} \cmidrule(lr){3-5} \cmidrule(lr){6-8} \cmidrule(lr){9-11}
         & mAP@C & mAP@C & mAP@P & mAP@A & mAP@C & mAP@P & mAP@A & mAP@C & mAP@P & mAP@A \\
         \midrule
         MMA \cite{cermelli2022modeling} & 37.5 & 12.9 & 31.7 & 21.9 & 4.3 & 8.2 & 6.9 & 3.5 & 4.1 & 3.9 \\
         IOR \cite{an2025ior} & 38.7 & 20.5 & 30.5 & 25.3 & 23.8 & 4.8 & 11.1 & 24.0 & 1.4 & 7.0 \\
         \midrule
         CL-CLIP (Ours) & \textbf{48.1} & \textbf{35.6} & \textbf{17.3} & \textbf{26.2} & \textbf{32.7} & \textbf{13.4} & \textbf{19.8} & \textbf{29.1} & \textbf{9.3} & \textbf{14.3} \\
         \bottomrule
    \end{tabular}
    }
\end{table}

\begin{figure}[htbp]
    \centering
    \includegraphics[width=\textwidth]{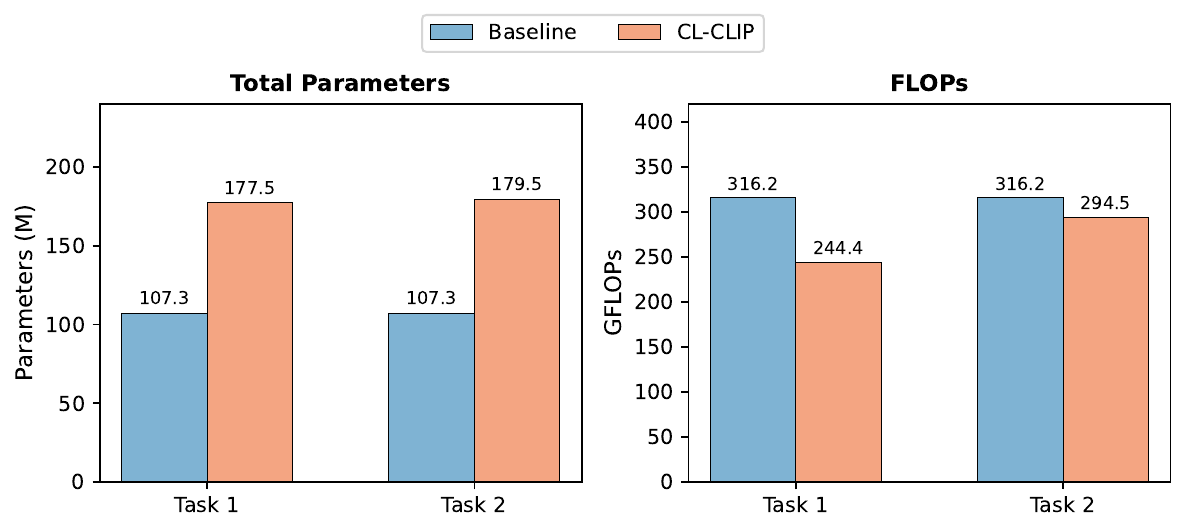}
    \caption{Inference efficiency on VOC 10+10. Params counts parameters activated during the forward pass. CL-CLIP achieves lower FLOPs than the baseline due to the lighter cost-volume pathway, with FLOPs growing moderately as more category experts become active.}
    \label{fig:efficiency}
\end{figure}

\paragraph{Efficiency analysis.}
Fig.~\ref{fig:efficiency} compares the inference cost of CL-CLIP and the F-ViT baseline on the VOC 10+10 setting. CL-CLIP has more total parameters (177.48M$\to$179.50M vs.\ baseline 107.26M) due to the per-category expert heads, with a slight increase as new experts are added. However, CL-CLIP achieves \emph{lower} FLOPs than the baseline (244.45\,G vs.\ 316.24\,G at Task\,1) because the cost-volume pathway is computationally lighter. As the number of seen categories grows, FLOPs increase (294.48\,G at Task\,2) but remain below the baseline in this 20-class setting. On larger-scale benchmarks with more categories (e.g., 80-class COCO), FLOPs grow proportionally with the number of seen-category experts, which could be addressed in future work through expert sharing or pruning strategies.

\paragraph{Multi-seed variance.}
To confirm reproducibility, we run CL-CLIP on the VOC 10+10 split with three random seeds. Tab.~\ref{tab:multi_seed} reports per-seed results and the mean$\pm$std. The low variance indicates that our method is stable across different initializations.

\begin{table}[htbp]
    \centering
    \caption{Multi-seed results on PASCAL VOC 10+10 (mAP@0.5)}
    \label{tab:multi_seed}
    \begin{tabular}{c c ccc}
         \toprule
         \multirow{2}{*}{\textbf{Seed}} & \textbf{Task1} & \multicolumn{3}{c}{\textbf{Task2}} \\
         \cmidrule(lr){2-2} \cmidrule(lr){3-5}
         & mAP@C & mAP@C & mAP@P & mAP@A \\
         \midrule
         0 & 86.2 & 84.5 & 64.8 & 74.7 \\
         1 & 86.4 & 84.8 & 63.7 & 74.3 \\
         2 & 86.4 & 84.7 & 64.9 & 74.8 \\
         \midrule
         Mean$\pm$Std & 86.3$\pm$0.1 & 84.7$\pm$0.2 & 64.5$\pm$0.7 & 74.6$\pm$0.3 \\
         \bottomrule
    \end{tabular}
\end{table}

%% file: checklist.tex
\section*{NeurIPS Paper Checklist}

\begin{enumerate}

\item {\bf Claims}
    \item[] Question: Do the main claims made in the abstract and introduction accurately reflect the paper's contributions and scope?
    \item[] Answer: \answerYes{} % Replace by \answerYes{}, \answerNo{}, or \answerNA{}.
    \item[] Justification: The abstract and introduction clearly state that CL-CLIP achieves competitive performance with existing continual object detectors and substantially improves the F-ViT baseline. These claims are directly supported by experimental results on PASCAL VOC and MS-COCO (Sec.~\ref{sec:experiments}). The paper also includes a Limitations paragraph discussing computational overhead of the per-expert design.
    \item[] Guidelines:
    \begin{itemize}
        \item The answer \answerNA{} means that the abstract and introduction do not include the claims made in the paper.
        \item The abstract and/or introduction should clearly state the claims made, including the contributions made in the paper and important assumptions and limitations. A \answerNo{} or \answerNA{} answer to this question will not be perceived well by the reviewers. 
        \item The claims made should match theoretical and experimental results, and reflect how much the results can be expected to generalize to other settings. 
        \item It is fine to include aspirational goals as motivation as long as it is clear that these goals are not attained by the paper. 
    \end{itemize}

\item {\bf Limitations}
    \item[] Question: Does the paper discuss the limitations of the work performed by the authors?
    \item[] Answer: \answerYes{} % Replace by \answerYes{}, \answerNo{}, or \answerNA{}.
    \item[] Justification: A dedicated Limitations paragraph at the end of Sec.~\ref{sec:experiments} discusses four aspects: (1)~single-seed evaluation without variance reporting, (2)~evaluation scope limited to VOC and COCO, (3)~dependence on CLIP text-image alignment quality, and (4)~linear growth of inference cost with the number of seen categories (detailed in Appendix.~\ref{Appendix E}).
    \item[] Guidelines:
    \begin{itemize}
        \item The answer \answerNA{} means that the paper has no limitation while the answer \answerNo{} means that the paper has limitations, but those are not discussed in the paper. 
        \item The authors are encouraged to create a separate ``Limitations'' section in their paper.
        \item The paper should point out any strong assumptions and how robust the results are to violations of these assumptions (e.g., independence assumptions, noiseless settings, model well-specification, asymptotic approximations only holding locally). The authors should reflect on how these assumptions might be violated in practice and what the implications would be.
        \item The authors should reflect on the scope of the claims made, e.g., if the approach was only tested on a few datasets or with a few runs. In general, empirical results often depend on implicit assumptions, which should be articulated.
        \item The authors should reflect on the factors that influence the performance of the approach. For example, a facial recognition algorithm may perform poorly when image resolution is low or images are taken in low lighting. Or a speech-to-text system might not be used reliably to provide closed captions for online lectures because it fails to handle technical jargon.
        \item The authors should discuss the computational efficiency of the proposed algorithms and how they scale with dataset size.
        \item If applicable, the authors should discuss possible limitations of their approach to address problems of privacy and fairness.
        \item While the authors might fear that complete honesty about limitations might be used by reviewers as grounds for rejection, a worse outcome might be that reviewers discover limitations that aren't acknowledged in the paper. The authors should use their best judgment and recognize that individual actions in favor of transparency play an important role in developing norms that preserve the integrity of the community. Reviewers will be specifically instructed to not penalize honesty concerning limitations.
    \end{itemize}

\item {\bf Theory assumptions and proofs}
    \item[] Question: For each theoretical result, does the paper provide the full set of assumptions and a complete (and correct) proof?
    \item[] Answer: \answerNA{} % Replace by \answerYes{}, \answerNo{}, or \answerNA{}.
    \item[] Justification: This paper does not include theoretical results; all contributions are empirical.
    \item[] Guidelines:
    \begin{itemize}
        \item The answer \answerNA{} means that the paper does not include theoretical results. 
        \item All the theorems, formulas, and proofs in the paper should be numbered and cross-referenced.
        \item All assumptions should be clearly stated or referenced in the statement of any theorems.
        \item The proofs can either appear in the main paper or the supplemental material, but if they appear in the supplemental material, the authors are encouraged to provide a short proof sketch to provide intuition. 
        \item Inversely, any informal proof provided in the core of the paper should be complemented by formal proofs provided in appendix or supplemental material.
        \item Theorems and Lemmas that the proof relies upon should be properly referenced. 
    \end{itemize}

    \item {\bf Experimental result reproducibility}
    \item[] Question: Does the paper fully disclose all the information needed to reproduce the main experimental results of the paper to the extent that it affects the main claims and/or conclusions of the paper (regardless of whether the code and data are provided or not)?
    \item[] Answer: \answerYes{} % Replace by \answerYes{}, \answerNo{}, or \answerNA{}.
    \item[] Justification: The full architecture is described with equations in Sec.~\ref{sec:experiments}. Appendix.~\ref{Appendix B} provides a detailed implementation table (optimizer, learning rate, batch size, training epochs, etc.) and the exact category splits for all task sequences on both MS-COCO and PASCAL VOC. The frozen/trainable boundary of each module is explicitly stated in the methodology.
    \item[] Guidelines:
    \begin{itemize}
        \item The answer \answerNA{} means that the paper does not include experiments.
        \item If the paper includes experiments, a \answerNo{} answer to this question will not be perceived well by the reviewers: Making the paper reproducible is important, regardless of whether the code and data are provided or not.
        \item If the contribution is a dataset and\slash or model, the authors should describe the steps taken to make their results reproducible or verifiable. 
        \item Depending on the contribution, reproducibility can be accomplished in various ways. For example, if the contribution is a novel architecture, describing the architecture fully might suffice, or if the contribution is a specific model and empirical evaluation, it may be necessary to either make it possible for others to replicate the model with the same dataset, or provide access to the model. In general. releasing code and data is often one good way to accomplish this, but reproducibility can also be provided via detailed instructions for how to replicate the results, access to a hosted model (e.g., in the case of a large language model), releasing of a model checkpoint, or other means that are appropriate to the research performed.
        \item While NeurIPS does not require releasing code, the conference does require all submissions to provide some reasonable avenue for reproducibility, which may depend on the nature of the contribution. For example
        \begin{enumerate}
            \item If the contribution is primarily a new algorithm, the paper should make it clear how to reproduce that algorithm.
            \item If the contribution is primarily a new model architecture, the paper should describe the architecture clearly and fully.
            \item If the contribution is a new model (e.g., a large language model), then there should either be a way to access this model for reproducing the results or a way to reproduce the model (e.g., with an open-source dataset or instructions for how to construct the dataset).
            \item We recognize that reproducibility may be tricky in some cases, in which case authors are welcome to describe the particular way they provide for reproducibility. In the case of closed-source models, it may be that access to the model is limited in some way (e.g., to registered users), but it should be possible for other researchers to have some path to reproducing or verifying the results.
        \end{enumerate}
    \end{itemize}

\item {\bf Open access to data and code}
    \item[] Question: Does the paper provide open access to the data and code, with sufficient instructions to faithfully reproduce the main experimental results, as described in supplemental material?
    \item[] Answer: \answerNo{} % Replace by \answerYes{}, \answerNo{}, or \answerNA{}.
    \item[] Justification: The code is not released at submission time as it is still being cleaned and organized. We will open-source the full codebase upon acceptance.
    \item[] Guidelines:
    \begin{itemize}
        \item The answer \answerNA{} means that paper does not include experiments requiring code.
        \item Please see the NeurIPS code and data submission guidelines (\url{https://neurips.cc/public/guides/CodeSubmissionPolicy}) for more details.
        \item While we encourage the release of code and data, we understand that this might not be possible, so \answerNo{} is an acceptable answer. Papers cannot be rejected simply for not including code, unless this is central to the contribution (e.g., for a new open-source benchmark).
        \item The instructions should contain the exact command and environment needed to run to reproduce the results. See the NeurIPS code and data submission guidelines (\url{https://neurips.cc/public/guides/CodeSubmissionPolicy}) for more details.
        \item The authors should provide instructions on data access and preparation, including how to access the raw data, preprocessed data, intermediate data, and generated data, etc.
        \item The authors should provide scripts to reproduce all experimental results for the new proposed method and baselines. If only a subset of experiments are reproducible, they should state which ones are omitted from the script and why.
        \item At submission time, to preserve anonymity, the authors should release anonymized versions (if applicable).
        \item Providing as much information as possible in supplemental material (appended to the paper) is recommended, but including URLs to data and code is permitted.
    \end{itemize}

\item {\bf Experimental setting/details}
    \item[] Question: Does the paper specify all the training and test details (e.g., data splits, hyperparameters, how they were chosen, type of optimizer) necessary to understand the results?
    \item[] Answer: \answerYes{} % Replace by \answerYes{}, \answerNo{}, or \answerNA{}.
    \item[] Justification: All training details (optimizer, learning rate schedule, batch size, number of epochs, input resolution, loss weights) and evaluation protocol (task splits, metrics) are specified in Appendix.~\ref{Appendix B}. Data splits for both MS-COCO and PASCAL VOC are listed explicitly.
    \item[] Guidelines:
    \begin{itemize}
        \item The answer \answerNA{} means that the paper does not include experiments.
        \item The experimental setting should be presented in the core of the paper to a level of detail that is necessary to appreciate the results and make sense of them.
        \item The full details can be provided either with the code, in appendix, or as supplemental material.
    \end{itemize}

\item {\bf Experiment statistical significance}
    \item[] Question: Does the paper report error bars suitably and correctly defined or other appropriate information about the statistical significance of the experiments?
    \item[] Answer: \answerYes{} % Replace by \answerYes{}, \answerNo{}, or \answerNA{}.
    \item[] Justification: We report mean$\pm$std over 3 random seeds on the VOC 10+10 setting (Tab.~\ref{tab:multi_seed} in Appendix~\ref{Appendix A}). The low variance confirms reproducibility of the main claims.
    \item[] Guidelines:
    \begin{itemize}
        \item The answer \answerNA{} means that the paper does not include experiments.
        \item The authors should answer \answerYes{} if the results are accompanied by error bars, confidence intervals, or statistical significance tests, at least for the experiments that support the main claims of the paper.
        \item The factors of variability that the error bars are capturing should be clearly stated (for example, train/test split, initialization, random drawing of some parameter, or overall run with given experimental conditions).
        \item The method for calculating the error bars should be explained (closed form formula, call to a library function, bootstrap, etc.)
        \item The assumptions made should be given (e.g., Normally distributed errors).
        \item It should be clear whether the error bar is the standard deviation or the standard error of the mean.
        \item It is OK to report 1-sigma error bars, but one should state it. The authors should preferably report a 2-sigma error bar than state that they have a 96\% CI, if the hypothesis of Normality of errors is not verified.
        \item For asymmetric distributions, the authors should be careful not to show in tables or figures symmetric error bars that would yield results that are out of range (e.g., negative error rates).
        \item If error bars are reported in tables or plots, the authors should explain in the text how they were calculated and reference the corresponding figures or tables in the text.
    \end{itemize}

\item {\bf Experiments compute resources}
    \item[] Question: For each experiment, does the paper provide sufficient information on the computer resources (type of compute workers, memory, time of execution) needed to reproduce the experiments?
    \item[] Answer: \answerYes{} % Replace by \answerYes{}, \answerNo{}, or \answerNA{}.
    \item[] Justification: All experiments are conducted on 4$\times$ NVIDIA RTX 4090 GPUs. Hardware details are reported in Appendix.~\ref{Appendix B}.
    \item[] Guidelines:
    \begin{itemize}
        \item The answer \answerNA{} means that the paper does not include experiments.
        \item The paper should indicate the type of compute workers CPU or GPU, internal cluster, or cloud provider, including relevant memory and storage.
        \item The paper should provide the amount of compute required for each of the individual experimental runs as well as estimate the total compute. 
        \item The paper should disclose whether the full research project required more compute than the experiments reported in the paper (e.g., preliminary or failed experiments that didn't make it into the paper). 
    \end{itemize}
    
\item {\bf Code of ethics}
    \item[] Question: Does the research conducted in the paper conform, in every respect, with the NeurIPS Code of Ethics \url{https://neurips.cc/public/EthicsGuidelines}?
    \item[] Answer: \answerYes{} % Replace by \answerYes{}, \answerNo{}, or \answerNA{}.
    \item[] Justification: This research uses only publicly available benchmark datasets (MS-COCO and PASCAL VOC) and does not involve human subject, private data, or potentially harmful applications.
    \item[] Guidelines:
    \begin{itemize}
        \item The answer \answerNA{} means that the authors have not reviewed the NeurIPS Code of Ethics.
        \item If the authors answer \answerNo, they should explain the special circumstances that require a deviation from the Code of Ethics.
        \item The authors should make sure to preserve anonymity (e.g., if there is a special consideration due to laws or regulations in their jurisdiction).
    \end{itemize}

\item {\bf Broader impacts}
    \item[] Question: Does the paper discuss both potential positive societal impacts and negative societal impacts of the work performed?
    \item[] Answer: \answerNA{} % Replace by \answerYes{}, \answerNo{}, or \answerNA{}.
    \item[] Justification: This work is foundational research on continual object detection using public benchmarks. It does not target any specific real-world deployment scenario and has no direct path to negative societal applications such as surveillance or discrimination.
    \item[] Guidelines:
    \begin{itemize}
        \item The answer \answerNA{} means that there is no societal impact of the work performed.
        \item If the authors answer \answerNA{} or \answerNo, they should explain why their work has no societal impact or why the paper does not address societal impact.
        \item Examples of negative societal impacts include potential malicious or unintended uses (e.g., disinformation, generating fake profiles, surveillance), fairness considerations (e.g., deployment of technologies that could make decisions that unfairly impact specific groups), privacy considerations, and security considerations.
        \item The conference expects that many papers will be foundational research and not tied to particular applications, let alone deployments. However, if there is a direct path to any negative applications, the authors should point it out. For example, it is legitimate to point out that an improvement in the quality of generative models could be used to generate Deepfakes for disinformation. On the other hand, it is not needed to point out that a generic algorithm for optimizing neural networks could enable people to train models that generate Deepfakes faster.
        \item The authors should consider possible harms that could arise when the technology is being used as intended and functioning correctly, harms that could arise when the technology is being used as intended but gives incorrect results, and harms following from (intentional or unintentional) misuse of the technology.
        \item If there are negative societal impacts, the authors could also discuss possible mitigation strategies (e.g., gated release of models, providing defenses in addition to attacks, mechanisms for monitoring misuse, mechanisms to monitor how a system learns from feedback over time, improving the efficiency and accessibility of ML).
    \end{itemize}
    
\item {\bf Safeguards}
    \item[] Question: Does the paper describe safeguards that have been put in place for responsible release of data or models that have a high risk for misuse (e.g., pre-trained language models, image generators, or scraped datasets)?
    \item[] Answer: \answerNA{} % Replace by \answerYes{}, \answerNo{}, or \answerNA{}.
    \item[] Justification: This paper does not release pre-trained models or datasets that pose a high risk of misuse.
    \item[] Guidelines:
    \begin{itemize}
        \item The answer \answerNA{} means that the paper poses no such risks.
        \item Released models that have a high risk for misuse or dual-use should be released with necessary safeguards to allow for controlled use of the model, for example by requiring that users adhere to usage guidelines or restrictions to access the model or implementing safety filters. 
        \item Datasets that have been scraped from the Internet could pose safety risks. The authors should describe how they avoided releasing unsafe images.
        \item We recognize that providing effective safeguards is challenging, and many papers do not require this, but we encourage authors to take this into account and make a best faith effort.
    \end{itemize}

\item {\bf Licenses for existing assets}
    \item[] Question: Are the creators or original owners of assets (e.g., code, data, models), used in the paper, properly credited and are the license and terms of use explicitly mentioned and properly respected?
    \item[] Answer: \answerYes{} % Replace by \answerYes{}, \answerNo{}, or \answerNA{}.
    \item[] Justification: All datasets (MS-COCO, PASCAL VOC) and pre-trained models (CLIP) are properly cited with their original publications. MS-COCO is released under CC BY 4.0 license, and PASCAL VOC is publicly available for research use.
    \item[] Guidelines:
    \begin{itemize}
        \item The answer \answerNA{} means that the paper does not use existing assets.
        \item The authors should cite the original paper that produced the code package or dataset.
        \item The authors should state which version of the asset is used and, if possible, include a URL.
        \item The name of the license (e.g., CC-BY 4.0) should be included for each asset.
        \item For scraped data from a particular source (e.g., website), the copyright and terms of service of that source should be provided.
        \item If assets are released, the license, copyright information, and terms of use in the package should be provided. For popular datasets, \url{paperswithcode.com/datasets} has curated licenses for some datasets. Their licensing guide can help determine the license of a dataset.
        \item For existing datasets that are re-packaged, both the original license and the license of the derived asset (if it has changed) should be provided.
        \item If this information is not available online, the authors are encouraged to reach out to the asset's creators.
    \end{itemize}

\item {\bf New assets}
    \item[] Question: Are new assets introduced in the paper well documented and is the documentation provided alongside the assets?
    \item[] Answer: \answerNA{} % Replace by \answerYes{}, \answerNo{}, or \answerNA{}.
    \item[] Justification: This paper does not release new datasets, models, or code packages.
    \item[] Guidelines:
    \begin{itemize}
        \item The answer \answerNA{} means that the paper does not release new assets.
        \item Researchers should communicate the details of the dataset\slash code\slash model as part of their submissions via structured templates. This includes details about training, license, limitations, etc. 
        \item The paper should discuss whether and how consent was obtained from people whose asset is used.
        \item At submission time, remember to anonymize your assets (if applicable). You can either create an anonymized URL or include an anonymized zip file.
    \end{itemize}

\item {\bf Crowdsourcing and research with human subjects}
    \item[] Question: For crowdsourcing experiments and research with human subjects, does the paper include the full text of instructions given to participants and screenshots, if applicable, as well as details about compensation (if any)? 
    \item[] Answer: \answerNA{} % Replace by \answerYes{}, \answerNo{}, or \answerNA{}.
    \item[] Justification: This paper does not involve crowdsourcing nor research with human subjects.
    \item[] Guidelines:
    \begin{itemize}
        \item The answer \answerNA{} means that the paper does not involve crowdsourcing nor research with human subjects.
        \item Including this information in the supplemental material is fine, but if the main contribution of the paper involves human subjects, then as much detail as possible should be included in the main paper. 
        \item According to the NeurIPS Code of Ethics, workers involved in data collection, curation, or other labor should be paid at least the minimum wage in the country of the data collector. 
    \end{itemize}

\item {\bf Institutional review board (IRB) approvals or equivalent for research with human subjects}
    \item[] Question: Does the paper describe potential risks incurred by study participants, whether such risks were disclosed to the subjects, and whether Institutional Review Board (IRB) approvals (or an equivalent approval/review based on the requirements of your country or institution) were obtained?
    \item[] Answer: \answerNA{} % Replace by \answerYes{}, \answerNo{}, or \answerNA{}.
    \item[] Justification: This paper does not involve crowdsourcing nor research with human subjects.
    \item[] Guidelines:
    \begin{itemize}
        \item The answer \answerNA{} means that the paper does not involve crowdsourcing nor research with human subjects.
        \item Depending on the country in which research is conducted, IRB approval (or equivalent) may be required for any human subjects research. If you obtained IRB approval, you should clearly state this in the paper. 
        \item We recognize that the procedures for this may vary significantly between institutions and locations, and we expect authors to adhere to the NeurIPS Code of Ethics and the guidelines for their institution. 
        \item For initial submissions, do not include any information that would break anonymity (if applicable), such as the institution conducting the review.
    \end{itemize}

\item {\bf Declaration of LLM usage}
    \item[] Question: Does the paper describe the usage of LLMs if it is an important, original, or non-standard component of the core methods in this research? Note that if the LLM is used only for writing, editing, or formatting purposes and does \emph{not} impact the core methodology, scientific rigor, or originality of the research, declaration is not required.
    %this research? 
    \item[] Answer: \answerNA{} % Replace by \answerYes{}, \answerNo{}, or \answerNA{}.
    \item[] Justification: The core methodology does not involve LLMs as any component. LLMs were only-used for writing assistance, which does not require declaration per NeurIPS policy.
    \item[] Guidelines:
    \begin{itemize}
        \item The answer \answerNA{} means that the core method development in this research does not involve LLMs as any important, original, or non-standard components.
        \item Please refer to our LLM policy in the NeurIPS handbook for what should or should not be described.
    \end{itemize}

\end{enumerate}

%% file: neurips_2026.bbl
\begin{thebibliography}{39}
\providecommand{\natexlab}[1]{#1}
\providecommand{\url}[1]{\texttt{#1}}
\expandafter\ifx\csname urlstyle\endcsname\relax
  \providecommand{\doi}[1]{doi: #1}\else
  \providecommand{\doi}{doi: \begingroup \urlstyle{rm}\Url}\fi

\bibitem[An et~al.(2025)An, Diao, Huang, Liu, An, and Xu]{an2025ior}
Zijia An, Boyu Diao, Libo Huang, Ruiqi Liu, Zhulin An, and Yongjun Xu.
\newblock Ior: Inversed objects replay for incremental object detection.
\newblock In \emph{ICASSP 2025-2025 IEEE International Conference on Acoustics, Speech and Signal Processing (ICASSP)}, pages 1--5. IEEE, 2025.

\bibitem[Bhatt et~al.(2024)Bhatt, Ross, and Sigal]{bhatt2024preventing}
Gaurav Bhatt, James Ross, and Leonid Sigal.
\newblock Preventing catastrophic forgetting through memory networks in continuous detection.
\newblock In \emph{European Conference on Computer Vision}, pages 442--458. Springer, 2024.

\bibitem[Carion et~al.(2020)Carion, Massa, Synnaeve, Usunier, Kirillov, and Zagoruyko]{carion2020end}
Nicolas Carion, Francisco Massa, Gabriel Synnaeve, Nicolas Usunier, Alexander Kirillov, and Sergey Zagoruyko.
\newblock End-to-end object detection with transformers.
\newblock In \emph{European conference on computer vision}, pages 213--229. Springer, 2020.

\bibitem[Cermelli et~al.(2022)Cermelli, Geraci, Fontanel, and Caputo]{cermelli2022modeling}
Fabio Cermelli, Antonino Geraci, Dario Fontanel, and Barbara Caputo.
\newblock Modeling missing annotations for incremental learning in object detection.
\newblock In \emph{Proceedings of the IEEE/CVF conference on computer vision and pattern recognition}, pages 3700--3710, 2022.

\bibitem[Chen et~al.(2019)Chen, Wang, Pang, Cao, Xiong, Li, Sun, Feng, Liu, Xu, et~al.]{chen1906mmdetection}
Kai Chen, Jiaqi Wang, Jiangmiao Pang, Yuhang Cao, Yu~Xiong, Xiaoxiao Li, Shuyang Sun, Wansen Feng, Ziwei Liu, Jiarui Xu, et~al.
\newblock Mmdetection: Open mmlab detection toolbox and benchmark. arxiv 2019.
\newblock \emph{arXiv preprint arXiv:1906.07155}, 5, 2019.

\bibitem[Cho et~al.(2024)Cho, Shin, Hong, Arnab, Seo, and Kim]{cho2024cat}
Seokju Cho, Heeseong Shin, Sunghwan Hong, Anurag Arnab, Paul~Hongsuck Seo, and Seungryong Kim.
\newblock Cat-seg: Cost aggregation for open-vocabulary semantic segmentation.
\newblock In \emph{Proceedings of the IEEE/CVF Conference on Computer Vision and Pattern Recognition}, pages 4113--4123, 2024.

\bibitem[Everingham et~al.(2010)Everingham, Van~Gool, Williams, Winn, and Zisserman]{everingham2010pascal}
Mark Everingham, Luc Van~Gool, Christopher~KI Williams, John Winn, and Andrew Zisserman.
\newblock The pascal visual object classes (voc) challenge.
\newblock \emph{International journal of computer vision}, 88\penalty0 (2):\penalty0 303--338, 2010.

\bibitem[Feng et~al.(2022)Feng, Wang, and Yuan]{feng2022overcoming}
Tao Feng, Mang Wang, and Hangjie Yuan.
\newblock Overcoming catastrophic forgetting in incremental object detection via elastic response distillation.
\newblock In \emph{Proceedings of the IEEE/CVF conference on computer vision and pattern recognition}, pages 9427--9436, 2022.

\bibitem[Gupta et~al.(2022)Gupta, Narayan, Joseph, Khan, Khan, and Shah]{gupta2022ow}
Akshita Gupta, Sanath Narayan, KJ~Joseph, Salman Khan, Fahad~Shahbaz Khan, and Mubarak Shah.
\newblock Ow-detr: Open-world detection transformer.
\newblock In \emph{Proceedings of the IEEE/CVF conference on computer vision and pattern recognition}, pages 9235--9244, 2022.

\bibitem[Huang et~al.(2023)Huang, He, Liu, and Wang]{huang2023incremental}
Ziyue Huang, Yupeng He, Qingjie Liu, and Yunhong Wang.
\newblock Incremental object detection with clip.
\newblock \emph{arXiv preprint arXiv:2310.08815}, 2023.

\bibitem[Jing et~al.(2024)Jing, He, Luo, Fei, Yang, Wei, Zhao, and Lu]{jing2024fineclip}
Dong Jing, Xiaolong He, Yutian Luo, Nanyi Fei, Guoxing Yang, Wei Wei, Huiwen Zhao, and Zhiwu Lu.
\newblock Fineclip: Self-distilled region-based clip for better fine-grained understanding.
\newblock \emph{Advances in Neural Information Processing Systems}, 37:\penalty0 27896--27918, 2024.

\bibitem[Kirkpatrick et~al.(2017)Kirkpatrick, Pascanu, Rabinowitz, Veness, Desjardins, Rusu, Milan, Quan, Ramalho, Grabska-Barwinska, et~al.]{kirkpatrick2017overcoming}
James Kirkpatrick, Razvan Pascanu, Neil Rabinowitz, Joel Veness, Guillaume Desjardins, Andrei~A Rusu, Kieran Milan, John Quan, Tiago Ramalho, Agnieszka Grabska-Barwinska, et~al.
\newblock Overcoming catastrophic forgetting in neural networks.
\newblock \emph{Proceedings of the national academy of sciences}, 114\penalty0 (13):\penalty0 3521--3526, 2017.

\bibitem[Kuo et~al.(2022)Kuo, Cui, Gu, Piergiovanni, and Angelova]{kuo2022f}
Weicheng Kuo, Yin Cui, Xiuye Gu, AJ~Piergiovanni, and Anelia Angelova.
\newblock F-vlm: Open-vocabulary object detection upon frozen vision and language models.
\newblock \emph{arXiv preprint arXiv:2209.15639}, 2022.

\bibitem[Li et~al.(2022)Li, Zhang, Zhang, Yang, Li, Zhong, Wang, Yuan, Zhang, Hwang, et~al.]{li2022grounded}
Liunian~Harold Li, Pengchuan Zhang, Haotian Zhang, Jianwei Yang, Chunyuan Li, Yiwu Zhong, Lijuan Wang, Lu~Yuan, Lei Zhang, Jenq-Neng Hwang, et~al.
\newblock Grounded language-image pre-training.
\newblock In \emph{Proceedings of the IEEE/CVF conference on computer vision and pattern recognition}, pages 10965--10975, 2022.

\bibitem[Li and Hoiem(2017)]{li2017learning}
Zhizhong Li and Derek Hoiem.
\newblock Learning without forgetting.
\newblock \emph{IEEE transactions on pattern analysis and machine intelligence}, 40\penalty0 (12):\penalty0 2935--2947, 2017.

\bibitem[Lin et~al.(2014)Lin, Maire, Belongie, Hays, Perona, Ramanan, Doll{\'a}r, and Zitnick]{lin2014microsoft}
Tsung-Yi Lin, Michael Maire, Serge Belongie, James Hays, Pietro Perona, Deva Ramanan, Piotr Doll{\'a}r, and C~Lawrence Zitnick.
\newblock Microsoft coco: Common objects in context.
\newblock In \emph{European conference on computer vision}, pages 740--755. Springer, 2014.

\bibitem[Lin et~al.(2017)Lin, Doll{\'a}r, Girshick, He, Hariharan, and Belongie]{lin2017feature}
Tsung-Yi Lin, Piotr Doll{\'a}r, Ross Girshick, Kaiming He, Bharath Hariharan, and Serge Belongie.
\newblock Feature pyramid networks for object detection.
\newblock In \emph{Proceedings of the IEEE conference on computer vision and pattern recognition}, pages 2117--2125, 2017.

\bibitem[Liu et~al.(2020)Liu, Kuang, Chen, Xue, Yang, and Zhang]{liu2020incdet}
Liyang Liu, Zhanghui Kuang, Yimin Chen, Jing-Hao Xue, Wenming Yang, and Wayne Zhang.
\newblock Incdet: In defense of elastic weight consolidation for incremental object detection.
\newblock \emph{IEEE transactions on neural networks and learning systems}, 32\penalty0 (6):\penalty0 2306--2319, 2020.

\bibitem[Liu et~al.(2024)Liu, Zeng, Ren, Li, Zhang, Yang, Jiang, Li, Yang, Su, et~al.]{liu2024grounding}
Shilong Liu, Zhaoyang Zeng, Tianhe Ren, Feng Li, Hao Zhang, Jie Yang, Qing Jiang, Chunyuan Li, Jianwei Yang, Hang Su, et~al.
\newblock Grounding dino: Marrying dino with grounded pre-training for open-set object detection.
\newblock In \emph{European conference on computer vision}, pages 38--55. Springer, 2024.

\bibitem[Liu et~al.(2023)Liu, Cong, Goswami, Liu, and Van De~Weijer]{liu2023augmented}
Yuyang Liu, Yang Cong, Dipam Goswami, Xialei Liu, and Joost Van De~Weijer.
\newblock Augmented box replay: Overcoming foreground shift for incremental object detection.
\newblock In \emph{Proceedings of the IEEE/CVF international conference on computer vision}, pages 11367--11377, 2023.

\bibitem[Luo et~al.(2025)Luo, Zhang, Cheng, Xing, Liang, Wang, and Zhang]{luo2025gradient}
Wenlong Luo, Shizhou Zhang, De~Cheng, Yinghui Xing, Guoqiang Liang, Peng Wang, and Yanning Zhang.
\newblock Gradient decomposition and alignment for incremental object detection.
\newblock In \emph{Proceedings of the IEEE/CVF International Conference on Computer Vision}, pages 4486--4495, 2025.

\bibitem[Paszke et~al.(2019)Paszke, Gross, Massa, Lerer, Bradbury, Chanan, Killeen, Lin, Gimelshein, Antiga, et~al.]{paszke2019pytorch}
Adam Paszke, Sam Gross, Francisco Massa, Adam Lerer, James Bradbury, Gregory Chanan, Trevor Killeen, Zeming Lin, Natalia Gimelshein, Luca Antiga, et~al.
\newblock Pytorch: An imperative style, high-performance deep learning library.
\newblock \emph{Advances in neural information processing systems}, 32, 2019.

\bibitem[Peng et~al.(2020)Peng, Zhao, and Lovell]{peng2020faster}
Can Peng, Kun Zhao, and Brian~C Lovell.
\newblock Faster ilod: Incremental learning for object detectors based on faster rcnn.
\newblock \emph{Pattern recognition letters}, 140:\penalty0 109--115, 2020.

\bibitem[Radford et~al.(2021)Radford, Kim, Hallacy, Ramesh, Goh, Agarwal, Sastry, Askell, Mishkin, Clark, et~al.]{radford2021learning}
Alec Radford, Jong~Wook Kim, Chris Hallacy, Aditya Ramesh, Gabriel Goh, Sandhini Agarwal, Girish Sastry, Amanda Askell, Pamela Mishkin, Jack Clark, et~al.
\newblock Learning transferable visual models from natural language supervision.
\newblock In \emph{International conference on machine learning}, pages 8748--8763. PMLR, 2021.

\bibitem[Rebuffi et~al.(2017)Rebuffi, Kolesnikov, Sperl, and Lampert]{rebuffi2017icarl}
Sylvestre-Alvise Rebuffi, Alexander Kolesnikov, Georg Sperl, and Christoph~H Lampert.
\newblock icarl: Incremental classifier and representation learning.
\newblock In \emph{Proceedings of the IEEE conference on Computer Vision and Pattern Recognition}, pages 2001--2010, 2017.

\bibitem[Ren et~al.(2015)Ren, He, Girshick, and Sun]{ren2015faster}
Shaoqing Ren, Kaiming He, Ross Girshick, and Jian Sun.
\newblock Faster r-cnn: Towards real-time object detection with region proposal networks.
\newblock \emph{Advances in neural information processing systems}, 28, 2015.

\bibitem[Sun et~al.(2023)Sun, Fang, Wu, Wang, and Cao]{sun2023eva}
Quan Sun, Yuxin Fang, Ledell Wu, Xinlong Wang, and Yue Cao.
\newblock Eva-clip: Improved training techniques for clip at scale.
\newblock \emph{arXiv preprint arXiv:2303.15389}, 2023.

\bibitem[Tschannen et~al.(2025)Tschannen, Gritsenko, Wang, Naeem, Alabdulmohsin, Parthasarathy, Evans, Beyer, Xia, Mustafa, et~al.]{tschannen2025siglip}
Michael Tschannen, Alexey Gritsenko, Xiao Wang, Muhammad~Ferjad Naeem, Ibrahim Alabdulmohsin, Nikhil Parthasarathy, Talfan Evans, Lucas Beyer, Ye~Xia, Basil Mustafa, et~al.
\newblock Siglip 2: Multilingual vision-language encoders with improved semantic understanding.
\newblock \emph{Localization, and Dense Features}, 6, 2025.

\bibitem[Wang et~al.(2025)Wang, Wang, and Lin]{wang2025gcd}
Xu~Wang, Zilei Wang, and Zihan Lin.
\newblock Gcd: Advancing vision-language models for incremental object detection via global alignment and correspondence distillation.
\newblock In \emph{Proceedings of the AAAI Conference on Artificial Intelligence}, volume~39, pages 8015--8023, 2025.

\bibitem[Woo et~al.(2018)Woo, Park, Lee, and Kweon]{woo2018cbam}
Sanghyun Woo, Jongchan Park, Joon-Young Lee, and In~So Kweon.
\newblock Cbam: Convolutional block attention module.
\newblock In \emph{Proceedings of the European conference on computer vision (ECCV)}, pages 3--19, 2018.

\bibitem[Wu et~al.(2025)Wu, Zhang, Cheng, Xing, Xu, Wang, and Zhang]{wu2025demystifying}
Qirui Wu, Shizhou Zhang, De~Cheng, Yinghui Xing, Di~Xu, Peng Wang, and Yanning Zhang.
\newblock Demystifying catastrophic forgetting in two-stage incremental object detector.
\newblock \emph{arXiv preprint arXiv:2502.05540}, 2025.

\bibitem[Wu and He(2018)]{wu2018group}
Yuxin Wu and Kaiming He.
\newblock Group normalization.
\newblock In \emph{Proceedings of the European conference on computer vision (ECCV)}, pages 3--19, 2018.

\bibitem[Xie et~al.(2025)Xie, Wang, Kong, Li, Liang, Zhang, Leng, and Yin]{xie2025fg}
Chunyu Xie, Bin Wang, Fanjing Kong, Jincheng Li, Dawei Liang, Gengshen Zhang, Dawei Leng, and Yuhui Yin.
\newblock Fg-clip: Fine-grained visual and textual alignment.
\newblock \emph{arXiv preprint arXiv:2505.05071}, 2025.

\bibitem[Yang et~al.(2022)Yang, Deng, Shi, Li, Zhang, Xu, Zhao, Lin, and Liang]{yang2022continual}
Binbin Yang, Xinchi Deng, Han Shi, Changlin Li, Gengwei Zhang, Hang Xu, Shen Zhao, Liang Lin, and Xiaodan Liang.
\newblock Continual object detection via prototypical task correlation guided gating mechanism.
\newblock In \emph{Proceedings of the IEEE/CVF conference on computer vision and pattern recognition}, pages 9255--9264, 2022.

\bibitem[Yang et~al.(2023)Yang, Zhou, Hong, Zhang, Wei, Zeng, Qiao, and Wang]{yang2023pseudo}
Dongbao Yang, Yu~Zhou, Xiaopeng Hong, Aoting Zhang, Xin Wei, Linchengxi Zeng, Zhi Qiao, and Weipinng Wang.
\newblock Pseudo object replay and mining for incremental object detection.
\newblock In \emph{Proceedings of the 31st ACM International Conference on Multimedia}, pages 153--162, 2023.

\bibitem[Yi et~al.(2025)Yi, Xu, Qin, Chen, Wu, Li, and Lao]{yi2025idpa}
Huahui Yi, Wei Xu, Ziyuan Qin, Xi~Chen, Xiaohu Wu, Kang Li, and Qicheng Lao.
\newblock idpa: Instance decoupled prompt attention for incremental medical object detection.
\newblock \emph{arXiv preprint arXiv:2506.00406}, 2025.

\bibitem[Yin et~al.(2022)Yin, Perez-Rua, and Liang]{yin2022sylph}
Li~Yin, Juan~M Perez-Rua, and Kevin~J Liang.
\newblock Sylph: A hypernetwork framework for incremental few-shot object detection.
\newblock In \emph{Proceedings of the IEEE/CVF conference on computer vision and pattern recognition}, pages 9035--9045, 2022.

\bibitem[Zhang et~al.(2024)Zhang, Gao, Zeng, Tian, Tan, Zhang, Qu, Liu, and Xie]{zhang2024learning}
Hongquan Zhang, Bin-Bin Gao, Yi~Zeng, Xudong Tian, Xin Tan, Zhizhong Zhang, Yanyun Qu, Jun Liu, and Yuan Xie.
\newblock Learning task-aware language-image representation for class-incremental object detection.
\newblock In \emph{Proceedings of the AAAI Conference on Artificial Intelligence}, volume~38, pages 7096--7104, 2024.

\bibitem[Zhong et~al.(2022)Zhong, Yang, Zhang, Li, Codella, Li, Zhou, Dai, Yuan, Li, et~al.]{zhong2022regionclip}
Yiwu Zhong, Jianwei Yang, Pengchuan Zhang, Chunyuan Li, Noel Codella, Liunian~Harold Li, Luowei Zhou, Xiyang Dai, Lu~Yuan, Yin Li, et~al.
\newblock Regionclip: Region-based language-image pretraining.
\newblock In \emph{Proceedings of the IEEE/CVF conference on computer vision and pattern recognition}, pages 16793--16803, 2022.

\end{thebibliography}
